\documentclass{article}

\PassOptionsToPackage{numbers, sort&compress}{natbib}

\usepackage[preprint]{corl_2021}

\usepackage[utf8]{inputenc}
\usepackage[table,xcdraw]{xcolor}

\usepackage{booktabs}
\newcommand{\ra}[1]{\renewcommand{\arraystretch}{#1}}

\usepackage{amssymb}
\usepackage{bbm}
\usepackage{amsmath}
\usepackage{pgfplots}
\usepackage[normalem]{ulem}
\usepackage{dsfont}
\usepackage{verbatim}
\usepackage{xspace}
\usepackage{blindtext}
\usepackage{wrapfig}
\usepackage{float}

\usepackage{multirow}

\usepackage{titlesec}
\titlespacing{\paragraph} {0pt}{0.1ex plus 0.05ex minus .1ex}{1em}

\usepackage[font=small,labelfont=bf]{caption}

\usepackage{atbegshi,etoolbox}%

\newcommand{\discardpages}[1]{%
  \xdef\discard@pages{#1}%
  \AtBeginShipout{%
    \renewcommand*{\do}[1]{%
      \ifnum\value{page}=##1\relax%
        \AtBeginShipoutDiscard%
        \gdef\do####1{}%
      \fi%
    }%
    \expandafter\docsvlist\expandafter{\discard@pages}%
  }%
}
\newif\ifkeeppage
\newcommand{\keeppages}[1]{%
  \xdef\keep@pages{#1}%
  \AtBeginShipout{%
    \keeppagefalse%
    \renewcommand*{\do}[1]{%
      \ifnum\value{page}=##1\relax%
        \keeppagetrue%
        \gdef\do####1{}%
      \fi%
    }%
    \expandafter\docsvlist\expandafter{\keep@pages}%
    \ifkeeppage\else\AtBeginShipoutDiscard\fi%
  }%
}

\usepackage{arydshln}

\newcommand{\fns}[1]{#1}

\newcommand{\eat}[1]{}

\usepackage{algorithmicx} 
\usepackage[noend]{algpseudocode}
\usepackage{algorithm}
\usepackage[T1]{fontenc}

\algnewcommand\algorithmicdefinitions{\textbf{Definitions:}}
\algnewcommand\Definitions{\item[\algorithmicdefinitions]}
\renewcommand{\algorithmiccomment}[1]{{\color{gray}\raisebox{1px}{\texttt{\guillemotright}} #1}}

\algnewcommand{\LineComment}[1]{\Statex \hskip\ALG@thistlm \algorithmiccomment{#1}}
\algrenewcommand\alglinenumber[1]{\footnotesize #1:}
\algrenewcommand\algorithmicindent{1.0em}%

\algnewcommand\algorithmicswitch{\textbf{switch}}
\algnewcommand\algorithmiccase{\textbf{case}}
\algnewcommand\algorithmicdefault{\textbf{default}}
\algdef{SE}[SWITCH]{Switch}{EndSwitch}[1]{\algorithmicswitch\ #1\ \algorithmicdo}{\algorithmicend\ \algorithmicswitch}%
\algdef{SE}[CASE]{Case}{EndCase}[1]{\algorithmiccase\ #1}{\algorithmicend\ \algorithmiccase}%
\algdef{SE}[DEFAULT]{Default}{EndDefault}[1]{\algorithmicdefault\ #1}{\algorithmicend\ \algorithmicdefault}%
\algdef{SE}[DOWHILE]{Do}{doWhile}{\algorithmicdo}[1]{\algorithmicwhile\ #1}%
\algtext*{EndSwitch}%
\algtext*{EndCase}%
\algtext*{EndDefault}%
\algrenewcommand\algorithmicindent{0.6em}%

\makeatletter
\newcommand{\StatexIndent}[1][3]{%
  \setlength\@tempdima{\algorithmicindent}%
  \Statex\hskip\dimexpr#1\@tempdima\relax}

\DeclareMathOperator*{\E}{\mathbb{E}}

\newcommand{\nlstring}{\emph}

\newcommand{\alfred}{\textsc{ALFRED}\xspace}

\newcommand{\taskstr}{L}

\newcommand{\traj}{\Xi}

\newcommand{\action}{a}
\newcommand{\actiontype}{\texttt{type}}
\newcommand{\actionmask}{\texttt{mask}}

\newcommand{\state}{s}
\newcommand{\obs}{o}
\newcommand{\image}{I}
\newcommand{\stopaction}{a_{\rm STOP}}
\newcommand{\failaction}{a_{\rm FAIL}}
\newcommand{\passaction}{a_{\rm PASS}}

\newcommand{\allactions}{\mathcal{A}}
\newcommand{\allnavactions}{\mathcal{A}_{\rm nav}}
\newcommand{\allintactions}{\mathcal{A}_{\rm int}}
\newcommand{\allstates}{\mathcal{S}}

\newcommand{\dynamics}{\mathcal{T}}

\newcommand{\rotateleft}{\textsc{RotateLeft}}
\newcommand{\rotateright}{\textsc{RotateRight}}
\newcommand{\moveahead}{\textsc{MoveAhead}}
\newcommand{\lookdown}{\textsc{LookDown}}
\newcommand{\lookup}{\textsc{LookUp}}
\newcommand{\pickup}{\textsc{Pickup}}
\newcommand{\putobject}{\textsc{Put}}
\newcommand{\toggleon}{\textsc{ToggleOn}}
\newcommand{\toggleoff}{\textsc{ToggleOff}}
\newcommand{\open}{\textsc{Open}}
\newcommand{\close}{\textsc{Close}}
\newcommand{\slice}{\textsc{Slice}}

\newcommand{\modelname}{\textsc{HLSM}}

\newcommand{\policy}{\pi}
\newcommand{\hlpolicy}{\pi^{H}}
\newcommand{\llpolicy}{\pi^{L}}
\newcommand{\obsfunc}{F}
\newcommand{\staterepr}{\hat{s}}

\newcommand{\hlaction}{g}

\newcommand{\stopsubgoal}{\hlaction_{\rm STOP}}

\newcommand{\hlactiontype}{\texttt{type}}
\newcommand{\hlactionarg}{\texttt{arg}^{C}}
\newcommand{\hlactionmask}{\texttt{arg}^{M}}
\newcommand{\numactiontypes}{K}

\newcommand{\featurize}{\textsc{Afford}}
\newcommand{\lingunet}{\textsc{LingUNet}}
\newcommand{\maskrefiner}{\textsc{Refiner}}
\newcommand{\egotransform}{\textsc{EgoTransform}}

\newcommand{\pose}{P}

\newcommand{\pitch}{\omega_{p}}
\newcommand{\yaw}{\omega_{y}}

\newcommand{\semanticgrid}{V^{S}}
\newcommand{\obssemanticgrid}{\hat{V}^{S}}
\newcommand{\observedgrid}{V^{O}}
\newcommand{\currentobservedgrid}{\hat{V}^{O}}
\newcommand{\inventory}{v^{S}}

\newcommand{\egosegmentation}{I^{S}}
\newcommand{\depth}{I^{D}}

\newcommand{\taskemb}{\phi^{L}}
\newcommand{\staterepremb}{\phi^{s}}
\newcommand{\histemb}{\phi^{g}}
\newcommand{\histtensor}{\mathbf{H}}

\newcommand{\gotoskill}{\textsc{GoTo}}

\newcommand{\navmodel}{\textsc{NavModel}}

\usepackage{titlesec}
\titlespacing*{\section}
{0pt}{5pt plus 3pt minus 2pt}{4pt plus 2pt}
\titlespacing*{\subsection}
{0pt}{4pt plus 3pt minus 1pt}{3pt plus 2pt}
\titlespacing*{\paragraph}
{0pt}{2pt plus 1pt minus 1pt}{2pt plus 1pt}

\title{A Persistent Spatial Semantic Representation for \\ High-level Natural Language Instruction Execution}

\author{
Valts Blukis$^{1,2}$, 
Chris Paxton$^{1}$, 
Dieter Fox$^{1,3}$, 
Animesh Garg$^{1,4}$, 
Yoav Artzi$^{2}$ 
\\
$^{1}$NVIDIA \hspace{20pt}
$^{2}$Cornell University \hspace{20pt}
$^{3}$University of Washington \\
$^{4}$University of Toronto, Vector Institute
}

\begin{document}

\maketitle

\begin{abstract}
Natural language provides an accessible and expressive interface to specify long-term tasks for robotic agents. However, non-experts are likely to specify such tasks with high-level instructions, which abstract over specific robot actions through several layers of abstraction. We propose that key to bridging this gap between language and robot actions over long execution horizons are persistent representations. We propose a persistent spatial semantic representation method, and show how it enables building an agent that performs hierarchical reasoning to effectively execute long-term tasks. We evaluate our approach on the ALFRED benchmark and achieve state-of-the-art results, despite completely avoiding the commonly used step-by-step instructions. 
\url{https://hlsm-alfred.github.io/}

\end{abstract}

\section{Introduction}\label{sec:intro}
Mobile manipulation in a home environment requires addressing multiple challenges, including exploration and making long-term inference about actions to perform.
In addition to reasoning, robots require
an accessible, yet sufficiently expressive interface to specify their tasks.
Natural Language provides an intuitive mechanism for task specification, and coupled with advances in automated language understanding, is increasingly applied to embodied agents~\cite[e.g.,][]{tellex11grounding,matuszek2012learning,misra2014context,thomason2015learning,misra2017mapping,nyga2018grounding,blukis2018following,blukis2018mapping,blukis2019learning,patki2020language,blukis2020fewshot}.

In this paper, we study the problem of learning to map high-level natural language instructions  %
to low-level mobile manipulation actions in an interactive 3D environment~\cite{shridhar2020alfred}. 
Existing work largely studies language tightly aligned to the robot actions, either using single-sentence instructions~\cite[e.g.,][]{tellex11grounding,matuszek2012learning,misra2017mapping,blukis2019learning} or sequences of instructions~\cite{anderson2017vision, tan2019learning, anderson2020sim, abp2021leaderboard,pashevich2021episodic,lwit2021leaderboard}.
In contrast, we focus on high-level instructions, which provide more efficient human-robot communication, but require long-horizon reasoning across layers of abstraction to generate actions not explicitly specified in the instruction.

\begin{figure*}[t!]
\centering
\includegraphics[width=\textwidth]{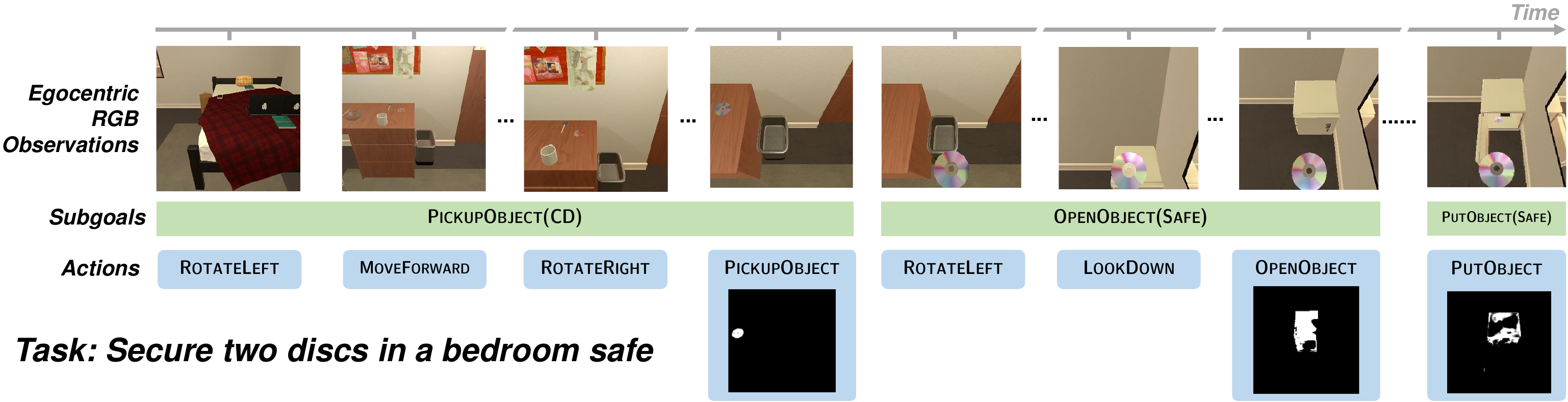}
\caption{Illustration of the task and our hierarchical formulation.
The agent receives a high-level task in natural language.
It needs to map RGB images to navigation and manipulation actions to complete the task.}
\label{fig:task}
\end{figure*}

Robust reasoning about manipulation goals from unrestricted high-level natural language instructions has a variety of open challenges. 
Consider the instruction \nlstring{secure two discs in a bedroom safe} (Figure~\ref{fig:task}).
The robot must first %
locate the \nlstring{safe} in the \nlstring{bedroom}. 
It then needs to distribute the actions entailed by \nlstring{secure} to two objects (\nlstring{two discs}), each requiring a distinct sequence of actions, but targeting the same \nlstring{safe}.
It is also required to map the verb \nlstring{secure} to its action space. %
In parallel, the robot must address mobile manipulation challenges, and often can only identify required actions as it observes and manipulates the world (e.g., if the \nlstring{safe} needs to be opened).

We propose to construct and continually update a spatial semantic representation of the world from robot observations (Figure~\ref{fig:modeloverview}).
Similar to widely used map representations~\cite{walter2013learning, hemachandra2015learning, patki2019inferring, kostavelis2015semantic},
we retain the spatial properties of the environment, allowing the robot to navigate and reason about relations between objects, as required to accomplish its task.
We propose the Hierarchical Language-conditioned Spatial Model (HLSM), a hierarchical approach that uses our spatial representation as a long-term memory to solve long-horizon tasks.
HLSM consists of a high-level controller that generates subgoals, and a low-level controller that generates sequences of actions to accomplish them.
In our example (Figure~\ref{fig:task}), the sequence of subgoals is $\langle$pick up a CD, open the safe, put the CD in the safe, \dots$\rangle$, each requiring a sequence of actions.
The spatial representation allows selecting subgoals that use previously observed objects outside of the agent's view, or to decide about needed exploration. 

We evaluate our approach on the \alfred~\cite{shridhar2020alfred} benchmark and achieve state-of-the-art results without using the low-level instructions used by previous work~\cite{abp2021leaderboard,pashevich2021episodic,lwit2021leaderboard,saha2021modular}, neither during training nor at test-time.
This paper makes three key contributions:
(a) a modular representation learning approach for the problem of mapping high-level natural language task descriptions to actions in a 3D environment;
(b) a method for utilizing a spatial semantic representation within a hierarchical model for solving mobile manipulation tasks; and (c) state-of-the-art performance on the \alfred benchmark, even outperforming all approaches that use detailed sequential instructions.

\section{Related Work}\label{sec:related}

Natural language has been extensively studied in robotics research, including with focus on instruction~\cite{tellex11grounding,kollar2010toward}, reference resolution~\cite{matuszek2012joint}, question generation~\cite{tellex2014asking, knepper15recovering, gong2018temporal}, and dialogue~\cite{brick2007incrementalnlp,tellexll2013toward,thomason2015learning}.
Most work in this area has considered either synthetic instructions of relatively simple goals~\cite{blukis2018following,hermann2017grounded, chaplot2017gated,paxton2019prospection}, or natural language instructions where all intermediate steps are explained in detail~\cite{misra2017mapping, anderson2017vision, suhr2018situated, fried2018speaker, misra2018mapping, ma2019selfmonitoring, tan2019learning, chen2019touchdown,shridhar2020alfred}. 
In contrast, we focus on high-level instructions, which are more likely in home environments~\cite{misra2015environment}. 

Representation of world state, action history, and language semantics plays a central role in robot systems and their algorithm design.
Symbolic representations have been extensively studied for instruction following agents~\cite{macmahon2006walk, branavan2010reading, tellex2011understanding, tellex11grounding, matuszek2012learning,duvallet2013imitation, walter2013learning, artzi2013weakly,misra2014context, thomason2015learning,misra2015environment,hemachandra2015learning, williams2018learning}.
While they simplify the symbol grounding problem and enable robustness, the ontologies on which they rely on are laborious to scale to new, unstructured  environments and language.
Representation learning presents an alternative by learning to map observations and language directly to actions~\cite{misra2017mapping, anderson2017vision, suhr2018situated,blukis2018mapping,blukis2019learning,blukis2020fewshot}.
World state and language semantics are represented with vectors~\cite{anderson2017vision} or by memorizing past observations~\cite{blukis2018mapping,pashevich2021episodic}.
Modelling improvements have enabled these approaches to achieve good performance on complex navigation tasks~\cite{anderson2017vision, blukis2018following, ma2019selfmonitoring, tan2019learning, blukis2019learning,blukis2020fewshot}, a success that has not yet translated to mobile manipulation~\cite{shridhar2020alfred, singh2020moca, zhang2021hitut}.

We propose integrating a semantic voxel map state representation within a hierarchical representation learning system.
Similar semantic 2D maps have been successfully used in navigation~\cite{gordon2017iqa, blukis2018following, blukis2018mapping, anderson2019chasing} and more recently even in mobile manipulation instruction-following tasks~\cite{saha2021modular}.
We extend these  maps to 3D and show state-of-the-art results on a challenging mobile manipulation benchmark.
Our map design is related to sparse metric, topological and semantic maps~\cite{walter2013learning, hemachandra2014learning, hemachandra2015learning, patki2019inferring, patki2020language} that have enabled grounding symbolic instruction representations.
Our map does not impose a topological structure or require reasoning about object instances, instead modelling a distribution over semantic classes for every voxel.

\section{Problem Definition}
\label{sec:task}

Let $\allactions$ be the set of agent actions, and $\allstates$ the set of world states.
Given a natural language instruction $\taskstr$ and an initial state $\state_{0} \in \allstates$, the agent's goal is to generate an execution $\traj = \langle \state_{0}, \action_{0}, \state_{1}, \action_{1}, \dots, \state_T, \action_T \rangle$, where $\action_{t} \in \allactions$ is an action taken by the agent at time $t$, $\state_{t} \in \allstates$ is the state before taking $\action_{t}$, and $\state_{t+1} = \dynamics(\state_{t}, \action_{t})$ under environment dynamics  $\dynamics : \allstates \times \allactions \rightarrow \allstates$.
The state $\state_{t}$ is defined by the environment layout and the poses and states of all objects and the agent.
The agent does not have access to the state $\state_{t}$, but only to an observation $\obs_{t}$.
An observation $\obs_{t} = (\image_{t}, \pose_{t}, \inventory_{t}, \taskstr)$ includes a first-person RGB camera image $\image_{t}$, the agent's pose $\pose_{t}$, a one-hot encoding of the object class the agent is holding $\inventory_{t}$, and the instruction $\taskstr$.%
The task is considered successful if all goal-conditions corresponding to the task $\taskstr$ are true at the final state $\state_T$.
Partial success is measured as the fraction of goal-conditions that have been achieved.

The \alfred dataset includes sets of seen and unseen environments.
The set of actions $\allactions = \allnavactions \cup \allintactions$ includes parameter-free navigation actions $\allnavactions = \{\moveahead, \rotateleft, \rotateright\}$ and interaction actions $\allintactions = \{\pickup, \putobject, \toggleon, \toggleoff, \open, \close, \slice\}$
parameterized by a binary mask that identifies the object of the interaction in the agent's current first-person view.
We compute $\pose_{t}$ and $\inventory_{t}$  using dead-reckoning from RGB observations and actions.

\begin{figure*}
\centering
    \includegraphics[width=0.8\textwidth]{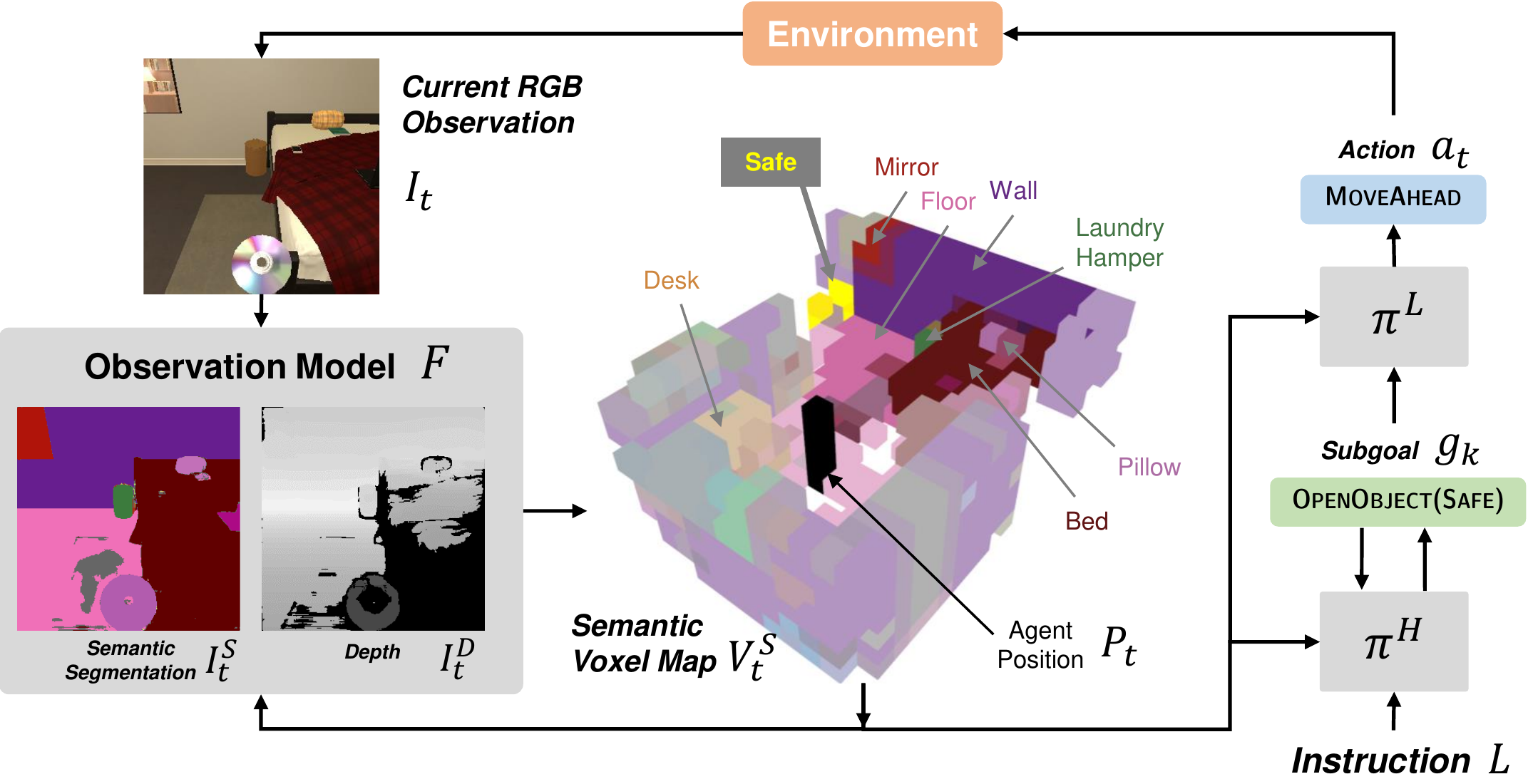}
    \captionof{figure}{Model architecture consisting of an observation model, high-level controller ($\hlpolicy$), and low-level controller ($\llpolicy$).
    The observation model updates the semantic voxel map state representation from RGB observations.
    $\hlpolicy$ predicts the next subgoal given the instruction and the map.
    $\llpolicy$ outputs a sequence of actions to achieve the subgoal.
    The semantic voxel map is visualized in the middle with agent position illustrated as a black pillar, ans the current sugoal argument mask
    in yellow.
    Other colors are different segmentation classes.
    Saturated voxels are observed in the current timestep.
    }
    \label{fig:modeloverview}
\end{figure*}

\section{Hierarchical Model with a Persistent Spatial Semantic Representation}\label{sec:model}

We model the agent behavior with a policy $\policy$ that maps an instruction $\taskstr$ and the observation $\obs_{t}$ at time $t$ to an action $\action_{t}$.
The policy $\policy$ is made of an \emph{observation model} $\obsfunc$ and two controllers: a \emph{high-level controller} $\hlpolicy$ and a \emph{low-level controller} $\llpolicy$.
The observation model builds a spatial \emph{state representation} $\staterepr_{t}$ that captures the cumulative agent knowledge of the world at time $t$.
$\staterepr_{t}$ is used by both $\hlpolicy$ for high-level long-horizon task planning, and  $\llpolicy$ for near-term reasoning, such as object search, navigation, collision avoidance, and manipulation.
Figure~\ref{fig:modeloverview} illustrates the policy.

The high-level controller $\hlpolicy$ computes a probability over \emph{subgoals}. 
A subgoal $\hlaction$ is a tuple $(\hlactiontype, \hlactionarg, \hlactionmask)$, where $\hlactiontype \in \allintactions$ is an interaction type (e.g., $\open$, $\pickup$), $\hlactionarg$ is the semantic class of the interaction argument (e.g., $\textsc{Safe}$, $\textsc{CD}$), and $\hlactionmask$ is a 3D mask identifying the location of the argument instance.
In \alfred, each interaction action in the set $\allintactions$ corresponds to a subgoal type.
When predicting the $k$-th subgoal at time $t$, $\hlpolicy$ considers the instruction $\taskstr$, the current state representation $\staterepr_{t}$, and the sequence of past subgoals $\langle \hlaction_i, \rangle_{i<k}$.
During inference, we sample from $\hlpolicy$. Unlike $\arg\max$, sampling allows the agent to re-try the same or different subgoal incase of a potentially random failure  (e.g., if a \textsc{Mug} was not found, pick up a \textsc{Cup}).

The low-level controller $\llpolicy$ is given the subgoal $\hlaction_{k}$ as its goal specification at time $t$.
At every timestep $j>t$, $\llpolicy$ maps the state representation $\staterepr_{j}$ and subgoal $\hlaction_{k}$ to an action $\action_{j}$, until it outputs one of the stop actions: $\passaction$ or $\failaction$ to indicate successful or failed subgoal completion.

The execution flow is as follows.
At time $t=0$ the initial observation $\obs_0$ is received.
At each timestep, we update the state representation $\staterepr_t$ using the observation model.
If there is no currently active subgoal, we sample a new subgoal $\hlaction_k$ from $\hlpolicy$, and then sample an action $\action_t$ from $\llpolicy$.
If $\action_t$ is $\passaction$, we increment subgoal counter $k$.
If it is $\failaction$, we discard the current subgoal $k$.
We repeat sampling subgoals and actions until an executable action $\action_t$ is sampled.
We execute $\action_t$, increment the timestep $t$, and receive the next observation $\obs_t$.
The episode ends when the subgoal $\stopsubgoal$ is sampled or the horizon $T_{max}$ is exceeded.
Algorithm~\ref{algo:exec} in Appendix~\ref{app:exec:flow} describes this process.

\subsection{State Representation}\label{sec:model:staterepr}

The state representation $\staterepr_t$ at time $t$ captures the agent's current understanding of the state of the world, including the locations of objects observed and the agent's relation to them.
The state representation is a tuple $(\semanticgrid_t, \observedgrid_t, \inventory_t, \pose_t)$.
The semantic map $\semanticgrid_{t} \in [0,1]^{X \times Y \times Z \times C}$
is a 3D voxel map that for every position
indicates which of the $c \in [1, C]$ object classes are present in the voxel.
The observability map $\observedgrid_t \in \{0, 1\}^{X \times Y \times Z}$ is a 3D voxel map that indicates whether the corresponding position has been observed.
The inventory vector $\inventory_t \in \{0, 1\}^C$   indicates which of the $C$ object classes the agent is currently holding.
The agent pose $\pose_t = (x, y, \pitch, \yaw)$ is specified by the 2D position $(x,y)$, pitch angle $\pitch$, and yaw angle $\yaw$.

We also compute 2D \emph{state affordance features} $\featurize(\staterepr_t) \in [0,1]^{7\times X\times Y}$ in a top-down view that represent each position with one or more of seven affordance classes  \{\texttt{pickable}, \texttt{receptacle}, \texttt{togglable}, \texttt{openable}, \texttt{ground}, \texttt{obstacle}, \texttt{observed}\}.
Each $[\featurize(\staterepr_t)]_{(\tau, x, y)} = 1.0$ if at least one of the voxels at position $(x,y)$ has affordance class $\tau$, otherwise it is zero.
$\featurize(\staterepr_t)$ is suited for object class agnostic reasoning, for example predicting a pose to pick up an object.
\footnote{We assume a known mapping between object semantic classes and affordance classes.}

\subsection{Observation Model}\label{sec:model:obs}

The observation model $\obsfunc(\staterepr_{t-1}, \obs_{t}, \hlaction_{k})$ updates the state representation with new observations. 
It considers the current subgoal $\hlaction_{k}$ to actively acquire information relevant to $\hlaction_{k}$.
The computation of $\obsfunc$ consists of three steps: perception, projection, accumulation.

\paragraph{Perception Step} We predict semantic segmentation $\egosegmentation_{t}$ and depth map $\depth_{t}$ from the RGB observation $\image_{t}$. 
We use neural networks pre-trained in the \alfred environment. 
The semantic segmentation $[\egosegmentation_{t}]_{(u,v)}$ is a distribution over $C$ object classes at pixel $(u,v)$.
The depth map $[\depth_{t}]_{(u,v)}$ is a binned distribution over $B$ bins.\footnote{We use $B$ uniformly spaced depth bins $\{0, \Delta_{D}, 2\Delta_{D}, \dots, (B-1)\Delta_{D} \}$, where $\Delta_{D}$ is a depth resolution. We suggest $\Delta_{D}$ should be less than 50\% of the voxel size. We used voxels with edge length 0.25m.} 
We also heuristically compute a binary mask $M^{D}_{t}$ that indicates which pixels have confident depth readings. We allow more confidence slack in pixels that correspond to the current subgoal argument $\hlactionarg_{t}$ according to $\egosegmentation_{t}$. Appendix~\ref{app:obsmodel} provides further details.
We use perception models based on the U-Net~\cite{ronneberger2015unet} architecture, but our framework supports other, potentially more powerful models as well (e.g. ~\cite{mccormac2017iccv, wu2019detectron2}).

\paragraph{Projection Step}
We use a pinhole camera model to convert depth $\depth_{t}$ and segmentation  $\egosegmentation_{t}$ to a point cloud that represents each image pixel $(u,v)$ with a 3D position $(x,y,z) \in \mathbbm{R}^{X \times Y \times Z}$ and a semantic distribution $[\egosegmentation_{t}]_{(u,v)}$.
We use $\arg \max_{B}(\depth_t)$ to compute the 3D positions, and  discard points at pixels $(u,v)$ when the binary mask value is $[M^{D}_{t}]_{(u,v)} = 0$. 
We construct a discrete semantic voxel map $\obssemanticgrid_t \in [0,1]^{X \times Y \times Z \times C}$, where $X$, $Y$, and $Z$ are the width, height, and length.
The value at each voxel $[\obssemanticgrid_t]_{(x,y,z)}$ is the element-wise maximum of the segmentation distributions $[\egosegmentation_{t}]_{(u,v)}$
across all points $(u,v)$ within the voxel.
We additionally compute a binary observability map $\currentobservedgrid_t \in \{0,1\}^{X \times Y \times Z}$ that indicates the voxels observed at time $t$.
A voxel is observed if it contains points, or if a ray cast from the camera through the voxel centroid has expected depth greater than the distance from the camera to the centroid.

\paragraph{Accumulation Step}

We integrate $\obssemanticgrid_t$ and $\currentobservedgrid_t$ into a persistent state representation:
\begin{align}
    \semanticgrid_{t} = \obssemanticgrid_t \times \currentobservedgrid_{t} + \semanticgrid_{t-1} \times (1 - \currentobservedgrid_{t})\hspace{1.2cm}\observedgrid_{t} = \max (\observedgrid_{t-1}, \currentobservedgrid_{t})\;\;.
\end{align}
This operation updates each voxel with the most recent semantic distribution, while retaining the values of all
voxels not visible at time $t$.
The output of the observation model is the spatial state representation $\staterepr_{t} = (\semanticgrid_t, \observedgrid_t, \inventory_t, \pose_t)$. 
The inventory $\inventory_t$ and pose $\pose_t$ are taken directly from 
$\obs_t$.

\subsection{High-level Controller ($\hlpolicy$)}\label{sec:model:hlp}

\begin{figure*}[t!]
\centering
\includegraphics[width=\textwidth]{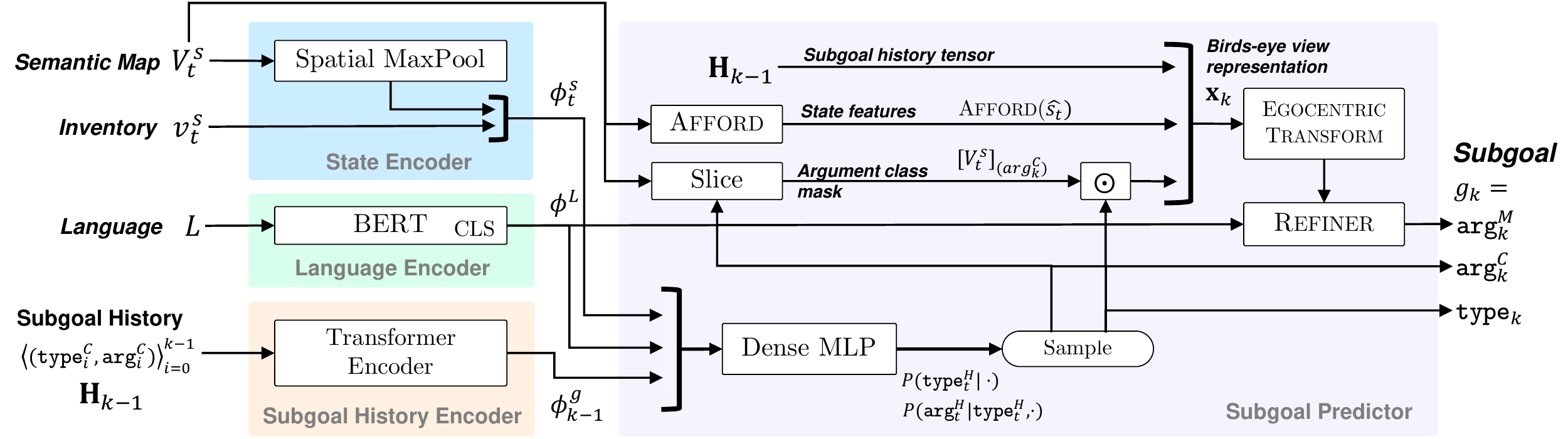}
\caption{Illustration of the high-level controller $\hlpolicy$ (Section~\ref{sec:model:hlp}).}
\label{fig:hlp}
\end{figure*}

At timestep $t$, when invoked for the $k$-th time, the input to $\hlpolicy$ is the instruction  $\taskstr$, the sequence of past subgoals $\langle \hlaction_i \rangle_{i<k}$,
and the current state representation $\staterepr_{t}$.
The output is the next subgoal $\hlaction_{k} = (\hlactiontype_{k}, \hlactionarg_{k}, \hlactionmask_{k})$. 
Figure~\ref{fig:hlp} illustrates the high-level controller architecture.

\paragraph{Input Encoding}
We encode the text $\taskstr$ using a pre-trained BERT~\cite{devlin2019bert} model that we fine-tune during training. We use the CLS token embedding as the task embedding $\taskemb$. 
We encode the state representation $\staterepr_t$  to account for classes of all observed objects, and the object that the agent is holding:  $\staterepremb(\staterepr_t) = [\inventory_t; \max_{(x,y,z)}(\semanticgrid_t)]$, where $\max_{(x,y,z)}$ is a max-pooling operation over spatial dimensions and $[\cdot ; \cdot]$ denotes concatenation.
We compute the representations of previous subgoals as $\langle \textsc{Repr}(\hlaction_{i}) \rangle_{i=0}^{k-1}$, where $\textsc{Repr}(\hlaction_{i})$ is the sum of a sinusoidal positional encoding~\cite{vaswani2017attention} of index $i$ and  learned embeddings  for $\hlactiontype_{i}$ and $\hlactionarg_{i}$.
We process this sequence with a two-layer Transformer autoregressive encoder~\cite{vaswani2017attention} to compute $\langle \histemb_i \rangle_{i=0}^{k-1}$.
We take $\histemb_{k-1}$ as the subgoal history embedding vector.
We additionally encode the argument mask information $\hlactionmask_{i}$ from the subgoal history in an integer-valued subgoal history tensor $\histtensor_{k-1} \in \mathbbm{N}^{\numactiontypes \times X\times Y}$ where $[\histtensor_{k-1}]_{(\tau,x,y)}$ is the number of times an interaction action type $\tau$ was performed at 2D position $(x,y)$ in the birds-eye view:
\vspace{-6pt}
\begin{equation}
\footnotesize
    [\histtensor_{k-1}]_{(\tau, x, y)} = \sum_{\substack{i = 0 \dots k-1 \\ \hlactionarg_i = \tau}}^{k-1}\max_{z}([\hlactionmask_{i}]_{(x,y,z)})\;\;.
\end{equation}
\vspace{-12pt}

\textbf{Subgoal Prediction}
We concatenate the three representations $\mathbf{h}_{(t, k)} = [\taskemb; \staterepremb_t; \histemb_{k-1}]$. We use a densely connected multi-layer perceptron~\cite{huang2017densely} to predict two distributions
$P(\hlactiontype_{k} \mid \mathbf{h}_{(t, k)})$ and $P(\hlactionarg_{k} \mid \hlactiontype_{k}, \mathbf{h}_{(t, k)})$, from which we sample a subgoal type $\hlactiontype_k$ and argument class $\hlactionarg_k$.

The remaining component of the subgoal is the action argument mask $\hlactionmask_{k}$. Let $[\semanticgrid_t]_{(\hlactionarg_k)}$ be a voxel map that only retains the object information for objects of class $\hlactionarg_k$ in the semantic map $\semanticgrid_{t}$.
We refine it to identify a single object instance. 
We compute a birds-eye view representation:%
\vspace{-0pt}
\begin{equation}
    \footnotesize
    \mathbf{x}_t = [\featurize(\staterepr_{t});\; \histtensor_{k-1};\; \max_z([\semanticgrid_t]_{(\hlactionarg_k)}) \otimes \mathbbm{1}_{\hlactiontype_k}]
\end{equation}
where $\featurize(\staterepr_{t})$ is a birds-eye view state affordance  feature map (Section~\ref{sec:model:staterepr}) and  $\mathbbm{1}_{\hlactiontype_k}$ is a one-hot encoding of $\hlactiontype_k$.\footnote{$\otimes$ denotes multiplication of a $X \times Y$ tensor with a $K$-dimensional vector to obtain a $K \times X \times Y$ tensor. $[\cdot;\cdot;\cdot]$ denotes channel-wise concatenation.}
Finally, we compute the 3D argument mask $\hlactionmask_k \in [0,1]^{X \times Y \times Z}$:
\begin{gather}
\footnotesize
    \hlactionmask_k = \maskrefiner(\egotransform(\mathbf{x}_t, \pose_{t}), \taskemb)\;\;,
    \label{eq:refiner}
\end{gather}
\vspace{-12pt}

where $\egotransform(\mathbf{x}, \pose_{t})$ transforms the map $\mathbf{x}$ to the agent egocentric pose $\pose_{t}$, 
$\maskrefiner$ is a neural network based on the LingUNet architecture~\cite{misra2018mapping}, and $\taskemb$ is the language embedding. 
The refined $\hlactionmask_{k}$ is a $[0,1]$-valued 3D mask that identifies the instance of the interaction argument object. If the object is believed to be unobserved, then $\hlactionmask_{k}$ contains all zeroes. The controller output is the subgoal $\hlaction_k = (\hlactiontype_k, \hlactionarg_k, \hlactionmask_k)$.

\subsection{Low-level Controller ($\llpolicy$)}

The low-level controller $\llpolicy$ is conditioned on the most recent subgoal $\hlaction_{k} = (\hlactiontype_k, \hlactionarg_k, \hlactionmask_k)$.
At time $t$, it maps the state representation $\staterepr_t$ to an action $\action_t$.  
It combines engineered and learned components. 
Appendix~\ref{app:model:llp} provides the implementation details. 
The controller $\llpolicy$ invokes a set of procedures: \texttt{NavigateTo}, \texttt{SampleExplorationPosition}, \texttt{SampleInteractionPose}, and \texttt{InteractMask}.
Their invokation follows a pre-specified execution flow across multiple timesteps.
First, we perform a 360\textdegree{} rotation to observe the nearby environment. 
If no objects of type $\hlactionarg_k$ are observed, we explore the environment by sampling a position $(x, y) = \texttt{SampleExplorationPosition}(\staterepr_{t})$, navigating there using the procedure $\texttt{NavigateTo}(x,y,\staterepr_{t})$, and performing a 360\textdegree{} rotation.
We repeat exploration until a voxel in $\semanticgrid_{t}$ contains the class $\hlactionarg_k$ with $>$50\% probability.
To interact with an object, we sample an interaction pose $(x, y, \yaw, \pitch) = \texttt{SampleInteractionPose}(\staterepr_{t}, \hlaction_{k})$, invoke $\texttt{NavigateTo}(x, y, \staterepr_{t})$ to reach the position (x,y), and then rotate according to yaw and pitch angles $(\yaw, \pitch)$.
Finally, we generate the egocentric interaction mask $\actionmask_{t} = \texttt{InteractMask}(\staterepr_{t}, \hlactionmask_k)$, and output the interaction action $(\hlactiontype_k, \actionmask_{t})$. 

All procedures use the spatial representation $\staterepr_{t}$.
\texttt{NavigateTo} navigates to a goal position using a value iteration network (VIN)~\cite{tamar2016valueiter} that reasons over obstacle and observability maps from $\staterepr_{t}$.
\texttt{SampleExplorationPosition} samples positions on the boundary of observed space in $\staterepr_{t}$.
\texttt{SampleInteractionPose} uses a learned neural network  $\navmodel$ to predict a distributon of poses from which the interaction $\hlaction_k$ will likely succeed.
\texttt{InteractMask} uses the segmentation image $\egosegmentation_{t}$ and the 3D argument mask $\hlactionmask_{t}$ to compute the first-person mask of the target object.

\section{Learning}\label{sec:learning}

The policy contains four learned models: the segmentation and depth networks, $\hlpolicy$, and the navigation model $\navmodel$ used by $\llpolicy$.
We train all four networks independently using supervised learning. 
We assume access to a training dataset \fns{$\mathcal{D} = \{(\taskstr^{(j)}, \traj^{(j)})\}_{j=1}^{N_{D}}$} of high-level natural language instructions \fns{$\taskstr^{(j)}$} paired with  demonstration execution \fns{$\traj^{(j)}$} in a set of seen environments.
Each execution \fns{$\traj^{(j)}$} is a sequence of states and actions \fns{$\langle \state_{0}^{(j)}, \action_{0}^{(j)}, \dots, \state_T^{(j)}, \action_T^{(j)} \rangle$}.
We denote $N_{P}$ the total number of states in dataset $\mathcal{D}$, and $N_{G}$ the total number of subgoals. 

We process \fns{$\mathcal{D}$} into three datasets.
The perception dataset \fns{$\mathcal{D}^{P} = \{([\image]^{(i)}, [\depth]^{(i)}, [\egosegmentation]^{(i)}\}_{i=1}^{N_{P}}$} includes RGB images \fns{$[\image]^{(i)}$} with ground truth depth \fns{$[\depth]^{(i)}$} and segmentation \fns{$[\egosegmentation]^{(i)}$}.
The subgoal dataset \fns{$\mathcal{D}^{g} = \{(\taskstr^{(i)}, \staterepr_{t}^{(i)}, \langle \hlaction_{j}^{(i)}\rangle_{j = 0}^{k})\}_{i=1}^{N_{G}}$} contains natural language instructions \fns{$\taskstr^{(i)}$}, state representations \fns{$\staterepr_{t}^{(i)}$} at the start of $k$-th subgoal execution, and sequences of the first $k$ subgoals \fns{$\langle \hlaction_{j}^{(i)}\rangle_{j = 0}^{k}$} extracted from \fns{$\traj^{(j)}$}.
The navigation dataset \fns{$\mathcal{D}^{N} = \{(\staterepr^{(i)},\hlaction^{(i)}, \pose^{(i)})\}_{i=1}^{N_{P}}$} consists of state representations \fns{$\staterepr^{(i)}$}, subgoals $\hlaction^{(i)}$, and agent poses \fns{$\pose^{(i)}$} at the time of taking the interaction action corresponding to subgoal \fns{$\hlaction^{(i)}$}.
The state representations \fns{$\staterepr^{(\cdot)}$} in datasets \fns{$\mathcal{D}^{g}$} and \fns{$\mathcal{D}^{N}$} are constructed using the observation model (Section~\ref{sec:model:obs}), but using ground-truth depth and segmentation images.

We train the perception models on \fns{$\mathcal{D}^{P}$} and the $\hlpolicy$ on \fns{$\mathcal{D}^{g}$} to predict the $k$-th subgoal by optimizing cross-entropy losses.
We use \fns{$\mathcal{D}^{N}$} to train the navigation model $\navmodel$ by optimizing a cross-entropy loss for positions and yaw angles, and an L2 loss for the pitch angle.

\section{Experimental Setup}\label{sec:experments}

\paragraph{Environment, Data, and Evaluation}
We evaluate our approach on the \alfred~\cite{shridhar2020alfred} benchmark.
It contains 108 training scenes, 88/4 validation seen/unseen scenes, and 107/8 test seen/unseen scenes.
There are 21{,}023 training tasks, 820/821 validation seen/unseen tasks, and 1533/1529 test seen/unseen tasks.
Each task is specified with a high-level natural language instruction.
The goal of the agent is to map raw RGB observations to actions to complete the task.
\alfred also provides detailed low-level step-by-step instructions, which simplify the reasoning process. We do not use these instructions for training or evaluation. 
We collect a training dataset of language-demonstration pairs for learning (Section~\ref{sec:learning}).
To extract subgoal sequences, we label each interaction action $\action_t = (\actiontype_t, \actionmask_t)$ and any preceding navigation actions with a single subgoal of $\hlactiontype = \actiontype_t$.
We compute the subgoal argument class $\hlactionarg$ and 3D mask $\hlactionmask$ labels from the first-person mask $\actionmask_t$, and ground truth segmentation and depth.
Completing a task requires satisfying several goal conditions. 
Following the common evaluation~\cite{shridhar2018interactive,alfredleaderboard}, we report two metrics.
\emph{Success rate} (SR) is the fraction of tasks for which all goal conditions were satisfied. \emph{Goal condition rate} (GC) is the fraction of goal-conditions satisfied across all tasks.

\paragraph{Systems}
We compare our approach, the Hierarchical Language-conditioned Spatial Model (HLSM)
to others on the ALFRED leaderboard that only use the high-level instructions.
At the time of writing, the only such published approach is HiTUT~\cite{zhang2021hitut}, an approach that uses a flat BERT~\cite{devlin2019bert} architecture to model a hierarchical task structure without using a spatial representation.
See Appendix~\ref{app:comp:hitut} for a detailed comparison.
We also compare to approaches that use the step-by-step instructions, which puts our method at a disadvantage.
Of these, LAV~\cite{nottingham2021lav} also imposes a hierarchical task structure and uses pre-trained depth and segmentation models, but without using a spatial state representation.

Additionally, we perform ablations and study sensory oracles.
To study the observation model, we compare to using sensory oracles for ground truth depth, ground truth segmentation, and both.
We report high-level controller ablations that remove the subgoal encoder, language encoder, and state representation encoder as used for predicting subgoal type $\hlactiontype_k$ and argument class $\hlactionarg_k$, while still using the state representation $\staterepr_t$ to predict the subgoal argument mask $\hlactionmask_k$.
We also study a low-level controller ablation that removes the exploration procedure.

\vspace{-5pt}
\section{Results}
\label{sec:results}
\vspace{-5pt}
\begin{figure*}[t!]
\centering
\includegraphics[width=\textwidth]{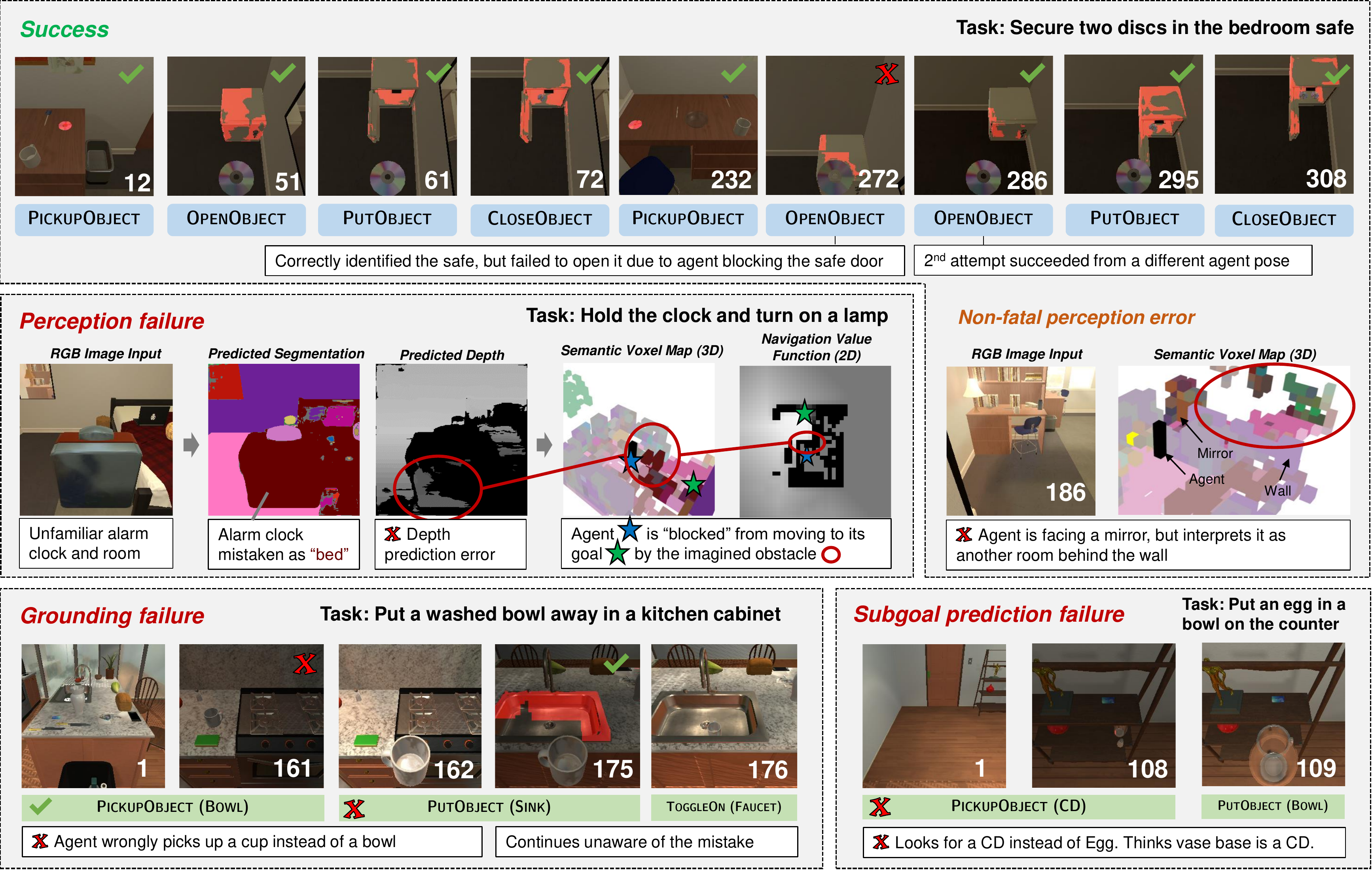}
\caption{Qualitative results showcasing successes and failures of our approach. \textbf{Top row:} snapshots of every interaction action taken during a successful task. Action argument masks are overlaid in red over the RGB images. The white numbers are timesteps.
\textbf{Middle-right}: illustration of a non-fatal perception error.
\textbf{Middle-left}: illustration of a fatal perception error.
The agent incorrectly interprets the reflection on the alarm clock as an obstacle, causing the agent (blue star) to believe that the path to the goal (green star) is blocked off.
This is reflected in the navigation value function computed by the value iteration network (VIN)~\cite{tamar2016valueiter}, where black cells are obstacles with value $-1$. White cell is the goal with value $1$.
\textbf{Bottom-left:} grounding failure. The agent wrongly picks up the cup instead of a bowl. Predicted subgoals are shown in green.
\textbf{Bottom-right:} high-level controller and percepton failure. $\hlpolicy$ predicts the wrong subgoal argument class (\textsc{CD} instead of \textsc{Egg}).
The segmentation model then mistakes the vase for a CD.
}
\label{fig:qualitative}
\end{figure*}

Table~\ref{tab:mainresults} shows test and validation results.
Our approach achieves state-of-the-art performance across both seen and unseen environments in the setting with only high-level instructions.
We achieve 10.04\% absolute (98.1\% relative) improvement in SR on the test unseen split, and 11.53\% absolute (62.6\% relative) improvement in SR on the test seen split compared to HiTUT G-only.

Our approach performs competitively even when compared to approaches that also use the low-level step-by-step instructions.
We achieve 4.84\% absolute (31.4\% relative) improvement in SR on the test unseen split compared to ABP~\cite{kim2021abp}.
On the test seen split, our approach performs reasonably well, however ABP~\cite{kim2021abp} and LWIT~\cite{lwit2021leaderboard} perform better, reflecting potentially stronger scene overfitting.

Tables~\ref{tab:ablations}~and~\ref{tab:pertaskresults} show development results.
We performed five runs of the full $\modelname$ model on the validation unseen data and found the sample standard deviation of the success rate is 1.1\% (absolute).
All other results are from a single-evaluation runs.
Ground truth depth alone (+ gt depth) does not significantly affect performance.
Ground truth segmentation (+ gt seg) provides 6.6\%/16.4\% absolute improvement in seen/unseen scenes. 
Using both (+ gt depth, gt seg) provides 11.1\%/21.9\% absolute improvement and narrows the seen/unseen gap from 11.3\% to 0.5\%.
This points to perception being the main bottleneck in generalization to unseen scenes.

We report high-level controller $\hlpolicy$ input encoder ablations.
The poor performance without the language encoder reflects task difficulty.
Zeroing the input to the subgoal history encoder (but keeping position encodings)
does not significantly affect performance,
showing that knowing the index of the current subgoal in addition to the state representation is often sufficient.
Not using the state representation for predicting subgoal type and argument class gives mixed results in seen and unseen scenes, but without a significant difference in performance.
Therefore, predicting the sequence of subgoal types and argument classes (i.e., what to do) is at times possible without spatial reasoning, while grounding the subgoal (i.e., where to do it) requires spatial information.
Removing random exploration from $\llpolicy$ does not significantly affect unseen performance.

Figure~\ref{fig:qualitative} illustrates the model behavior, showing both successes and common failures.
The main failures in valid unseen scenes are due to (1) perception errors that result in missing or extraneous obstacles or picking up wrong objects; (2) insufficiency of random exploration (e.g., not searching inside cabinets); (3) navigation model errors (e.g., blocking objects from opening); (4) subgoal prediction errors (e.g., picking up wrong objects); and (5) lack of state-aware multi-step planning and backtracking.
More qualitative results are available in Appendix~\ref{app:results}.

\newcommand{\rowspacing}{\ra{1.0}}
\newcommand{\rowsqueeze}{\vspace{-2pt}}

\newcommand{\bestH}{\textbf}
\newcommand{\bestF}{\underline}

\begin{table}
\centering
\footnotesize
\rowspacing
\begin{tabular}{@{}lcccccccccccc@{}}
\toprule
\multirow{3}{*}{\textbf{Method}}  && \multicolumn{5}{c}{\textbf{Test}} && \multicolumn{5}{c}{\textbf{Validation}} \rowsqueeze \\
 &&\multicolumn{2}{c}{Seen} && \multicolumn{2}{c}{Unseen} &&\multicolumn{2}{c}{Seen} && \multicolumn{2}{c}{Unseen} \rowsqueeze \\
 \cmidrule{3-4}\cmidrule{6-7}\cmidrule{9-10}\cmidrule{12-13}\rowsqueeze
  && SR & GC  && SR & GC && SR & GC  && SR & GC \\
\midrule
\multicolumn{13}{l}{\textbf{Low-level Sequential Instructions + High-level Goal Instruction}} \rowsqueeze\\
\midrule
\textsc{Seq2Seq}~\cite{shridhar2020alfred}    
             && 3.98    & 9.42  && 0.39   & 7.03    %
                  && 3.70    & 10.00 && 0.00   & 6.90 \\ %

\textsc{MOCA}~\cite{singh2020moca}    
                 && 22.05   & 28.29  && 5.30   & 14.28      %
                  && 19.15   & 28.5   && 3.78   & 13.4 \\    %
                  
\textsc{E.T.}~\cite{pashevich2021episodic}
                 && 28.77  & 36.47    && 5.04   & 15.01     %
                  && 33.78  & 42.48    && 3.17   & 13.12 \\  %
\textsc{E.T.} + synth. data~\cite{pashevich2021episodic}
                 && 38.42   & 45.44   && 8.57   & 18.6      %
                  && \bestF{46.59}   & \bestF{52.82}   && 7.32   & 20.87 \\  %

\textsc{LWIT}~\cite{nguyen2021lwit}
                & & 30.92   & 45.44   && 9.42   & 20.91     %
                  && 33.70   & 43.10   && 9.70   & 23.10 \\                 %
\textsc{HiTUT}\cite{zhang2021hitut}
    && 21.27   & 29.97   && 13.87   & 20.31       %
                  && 25.24   & 34.85   && 12.44   & 23.71 \\    %
\textsc{ABP}~\cite{kim2021abp}
                && \bestF{44.55}   & \bestF{51.13}   && 15.43   & 24.76       %
                  && 42.93   & 50.45   && 12.55   & 25.19 \\    %
\midrule
\multicolumn{13}{l}{\textbf{High-level Goal Instruction Only}} \rowsqueeze\\
\midrule
\textsc{HiTUT} G-only\cite{zhang2021hitut}
               && 18.41   & 25.27   && 10.23   & 20.27       %
                  && 13.63   & 21.11   && 11.12   & 17.89 \\    %
\textsc{LAV}~\cite{nottingham2021lav}
               && 13.35   & 23.21   && 6.38   & 17.27     %
                  && 12.7   & 23.4   && -   & - \\                 %

\textsc{\modelname} (Ours)    
             && \bestH{29.94}   & \bestH{41.21}   && \bestF{\bestH{20.27}}   & \bestF{\bestH{30.31}}       %
                  && \bestH{29.63}    & \bestH{38.74}    && \bestF{\bestH{18.28}}   & \bestF{\bestH{31.24}} \\    %
\bottomrule
\end{tabular}
\captionof{table}{Test results. Test seen/unseen and validation seen/unseen splits. Top section approaches use sequential step-by-step instructions. The bottom section uses only high-level instructions. Best results \textbf{using only high-level instructions} and \underline{using both types of instructions} are highlighted.}
\label{tab:mainresults}
\vspace{-10pt}
\end{table}

\begin{figure}
\begin{minipage}{.5\textwidth}
\centering
\footnotesize
\rowspacing
\begin{tabular}{@{}lccccc@{}}
\toprule
\multirow{3}{*}{\textbf{Method}} & \multicolumn{5}{c}{\textbf{Validation}} \rowsqueeze\\
   &\multicolumn{2}{c}{Seen} && \multicolumn{2}{c}{Unseen} \rowsqueeze\\
   \cmidrule{2-3}  \cmidrule{5-6} \rowsqueeze
   & SR & GC  && SR & GC \\
\midrule
$\modelname$
                & 29.6   & 38.8  & & 18.3   & \textbf{31.2} \\    %
\midrule
\;+ gt depth
                & 29.6   & 40.5   && 20.1   & 33.7 \\    %
\;+ gt depth, gt seg.
                & 40.7   & 50.4   && 40.2   & 52.2 \\    %
\;+ gt seg.
                & 36.2  & 47.0   && 34.7   & 47.8 \\    %
\midrule

\;w/o language enc.
                & \phantom{0}0.9   & \phantom{0}8.6  && \phantom{0}0.2   & \phantom{0}7.5\\    %
\;w/o subg. hist. enc.
                & 29.4   & 38.5   && 16.6   & 29.2 \\    %
\;w/o state repr enc.
                & 30.0   & 40.6   && \textbf{18.9}   & 30.8 \\    %
\midrule

\;w/o exploration
                & \textbf{32.2}   & \textbf{42.4}   && 18.1  & 31.3 \\    %
\bottomrule
\end{tabular}
\captionof{table}{Development results on validation split. Performance of our full approach, with perception oracles, a perception ablation, $\hlpolicy$ ablations, and $\llpolicy$ ablations}
\label{tab:ablations}

\end{minipage}
\hspace{1.5em}
\begin{minipage}{.36\textwidth}
\vspace{-0.9em}
\centering
\footnotesize
\rowspacing
\begin{tabular}{@{}lccccc@{}}
\toprule
\multirow{3}{*}{\textbf{Task Type}} & \multicolumn{5}{c}{\textbf{Validation}} \\
   &\multicolumn{2}{c}{Seen} && \multicolumn{2}{c}{Unseen} \\
   \cmidrule{2-3} \cmidrule{5-6}
   & SR & GC  && SR & GC \\
\midrule
Overall
                & 29.6   & 38.7   && 18.3   & 31.2 \\    %
\midrule
Examine
                & 46.8   & 59.0   && 36.6   & 59.9 \\    %
Pick \& Place
                & 57.0   & 57.0   && 34.8   & 34.8 \\    %
Stack \& Place
                & 13.0   & 27.0   && \phantom{0}4.4   & 14.3 \\    %
Clean \& Place
                & 25.0   & 39.5   && 11.3   & 25.8 \\    %
Cool \& Place
                & 17.5   & 33.8   && 14.8   & 39.6 \\    %
Heat \& Place
                & \phantom{0}9.3   & 29.1   && \phantom{0}0.0   & 17.0 \\    %
Pick 2 \& Place
                & 34.7   & 51.9   && 18.0   & 34.7 \\    %
\bottomrule
\end{tabular}
\vspace{0.5em}
\captionof{table}{Performance breakdown per task type on the validation split.}
\label{tab:pertaskresults}
\end{minipage}
\end{figure}

\section{Discussion and Limitations}
\label{sec:discussion}

We showed that a persistent spatial semantic representation enables a hierarchical model to achieve state-of-the-art performance on a challenging instruction-following mobile manipulation task.
The main performance bottlenecks include long-horizon exploration, perception generalization to unseen environments, and low-level motion planning for continuous collision avoidance.
In terms of learning, incorporating reinforcement learning to train $\hlpolicy$, $\llpolicy$, and observation model $\obsfunc$ jointly could improve robustness.
We defined the interface to $\llpolicy$ to be faithful to skills available on physical robots, but the exact implementation of $\llpolicy$ is not the focus of our work.
Physical deployment would require changes to $\llpolicy$, and study on robustness to errors in continuous environments, such as localization or motion uncertainty.

\clearpage
\newpage
\section{Acknowledgements}

This research was supported by ARO W911NF-21-1-0106, a Google Focused Award, and NSF under grant No. 1750499. 
Animesh Garg is supported in part by CIFAR AI Chair and NSERC Discovery Grant. 
A significant part of the work was done during the first author's internship at Nvidia.  
We thank the authors of ALFRED for maintaining the benchmark. We thank Mohit Shridhar and Jesse Thomason for their help answering our questions, and the anonymous reviewers for their helpful comments.

\bibliography{references}

\clearpage
\newpage
\appendix

\section{Appendix}

\subsection{Frequently Asked Questions}
\begin{itemize}
    \item \textbf{Are the \alfred sequential instructions needed during training?} The sequential step-by-step instructions are not needed neither during training, nor at test-time.

    \item \textbf{What has to be done to apply this approach to a real robot?}
    The observation model, high-level controller, state representation, and the interface to the low-level controller together constitute our contribution and are intended to generalize to physical robots.
    Deployment on a real robot would require an implementation of the low-level controller designed for continuous motion in cluttered environments, and an implementation of the \alfred interface to enable execution of manipulation actions such as $\pickup$ and $\toggleon$.
    Such physical robot capabilities are subject of  ongoing research~\cite{chiang2019learning, sundermeyer2021contact}.

    \item \textbf{Does this simulated environment result constitute progress towards real-world capabilities?}
    Real-robot operation is the long-term motivation of this work and has been carefully considered in the design of the representation and the approach. However, we do not claim to execute high-level natural language mobile manipulation instructions from raw vision on real robots in unseen environments. To date, such capabilities haven’t been demonstrated even in simulated environments, such as \alfred. Even in this scenario, though our method achieves better results than existing work, it can still only solve 18.28\% of problems in unseen environments.

    \item \textbf{Would the system scale to physically larger environments?} The main bottleneck towards scaling to larger environments is the memory constraint of the semantic memory.
    While our implementation is likely restricted to interior scenes when using commodity hardware, follow-up work could address this, perhaps using multi-scale representations such as Octress~\cite{meagher1982geometric, takikawa2021neural}.
    
    \item \textbf{How are the state dynamics modeled? Are they assumed to be known or are they learned?}
    The \gotoskill{} procedure in the low-level controller is based on a value-iteration network that utilizes a deterministic grid-navigation dynamics model on the internal representation, which is a crude approximation of the dynamics of the \rotateleft, \rotateright, \moveahead{} navigation commands. Other than that, the dynamics of the environment are assumed to be completely unknown to the agent, and are not explicitly learned or modeled.
    
    \item \textbf{How would localization uncertainty affect the approach?} Our representation approach assumes a reliable robot pose estimate.
    Precisely studying the effects of pose errors would require integration into a system for continuous environments.
    Intuitively, voxels further away are affected by pose errors more, but may better tolerate it due to being used mainly to decide navigation goals.
    Voxels close to the agent require more precision as they are used for object instance mask generation, but would be less affected by pose errors.
    Our voxel map uses a relatively coarse 25cm resolution.
    
    \item \textbf{Which model was used to obtain test results?} The full HLSM model was evaluated on the test set, even though the model without state representation encoding input to the high-level controller performed better in unseen environments on the validation set.
    
    \item \textbf{Why does the ablation without state representation encodings perform better in unseen environments?}
    In unseen environments, the semantic segmentation is erroneous due to the generalization gap, resulting in state encodings that contain errors.
    This may affect perfromance of the high-level controller that was trained on data with perfect segmentation, and thus with perfect state encodings.

    \item \textbf{What is the benefit of sampling the subgoals instead of attempting execution from most to least likely in order?}
    There are two types of subgoal execution failures: systematic and random. An example of a systematic failure is the selection of an incorrect subgoal. For example, \toggleon(\textsc{FloorLamp}) would fail if a \textsc{FloorLamp} does not exist in the environment. An example of a random failure is the low-level controller sampling an interaction pose for which the interaction fails (e.g., Figure 4, row 1, timestep 272).
    A next-best approach would alleviate a systematic failure, but a sampling approach alleviates both: the systematic failures by trying different subgoals, and random failures by potentially sampling the same subgoal multiple times.

\end{itemize}

\subsection{Extended Related Work}
\label{sec:relatedwork}

\paragraph{Grounding High-level Language to Actions in Robotics}

In order for natural language human-robot interfaces to be useful and widely adopted in practice, they should support  instructions that are as brief as possible while still being informative of the task, i.e., that adhere to Grice's maxim of quantity~\cite{grice1975logic}.
Following such high-level instructions requires bridging the gap from high-level language to long sequences of low-level actions.
This is commonly achieved using temporal abstraction, where subgoals or options abstract over sequences of low-level actions, reducing the effective time horizon of the problem. Most work on instruction following in robotics utilizes temporal abstraction~\cite{tellex11grounding, tellex2011understanding, artzi2013weakly, misra2014context, hemachandra2015learning, misra2015environment, misra2018mapping, paxton2019prospection, patki2019inferring, roh2020conditional, patki2020language}.

Various methods explicitly model correspondences between linguistic constituents in a symbolic instruction representation, environment percepts in the world model, and subgoals (behavior primitives)~\cite{kollar2010toward, tellex11grounding, matuszek2010following, misra2014context, hemachandra2015learning, misra2015environment}.
This requires the instruction to at least mention each subgoal, and precludes instructions that omit intermediate goals that are expected to be inferred. 
This limitation can be overcome by directly mapping from language to reward specifications~\cite{gopalan2018sequence, williams2018learning, bahdanau2018learning, goyal2020pixl2r} or post-conditions~\cite{misra2015environment}, and then using a planner~\cite{misra2015environment} or learning a task-specific policy~\cite{bahdanau2018learning, goyal2020pixl2r} to solve for the sequence of actions.
Both are difficult in practice.
Planning requires a compact, symbolic environment representation with an underlying ontology that is hard to construct for unstructured environments, such as the household environment studied in this work.
Policy learning is computationally expensive, and poorly adapts to novel tasks specified in natural language in real-time.

Recently, methods that map language and observations directly to actions using neural networks have seen rising popularity and success on simulated~\cite{anderson2017vision, blukis2018mapping, chen2019touchdown, tan2019learning, krantz2020beyond, jain2019stay, ku2020room} and real-robot~\cite{blukis2019learning, anderson2020sim, blukis2020fewshot} navigation, as well as simulated manipulation~\cite{misra2017mapping} tasks.
Simulated mobile manipulation is a promising next frontier~\cite{shridhar2020alfred, singh2020moca, zhang2021hitut}.
Representation learning approaches avoid planning, by using a direct sequence-to-sequence formulation and a data-driven approach that theoretically permits mapping arbitrarily terse input text to arbitrarily long action sequences that potentially include any necessary intermediate steps not explicitly mentioned in the text.
In practice, however, most research has focuses on relatively detailed step-by-step instructions, sometimes using modelling tools such as attention~\cite{anderson2017vision, pashevich2021episodic} and progress monitoring~\cite{ma2019selfmonitoring} to  leverage the sequential nature of the instructions.

We learn to follow high-level instructions in an interactive mobile manipulation 3D environment.
To bridge the gap between language and actions, we use temporal abstraction, where the high-level controller predicts subgoals that abstract over sequences of actions, and the low-level controller generates actions to fulfil each subgoal.
The controllers rely on a spatial-semantic state representation to enable reasoning about  what subgoals make progress towards the high-level task, and what actions make progress towards the specific subgoal, given all past sensory observations.
The persistent representation enables operation over long time horizons.
Using a shared world representation for both the high- and low-level controllers reduces representation engineering effort and error accumulation typically associated with pipeline approaches.

\paragraph{Semantic Maps for Language Grounding in Robotics}
The idea of building maps that combine spatial and semantic information~\cite{persson2007probabilistic, zender2008conceptual, pronobis2010multi, pronobis2011semantic, walter2013learning, hemachandra2014learning} and using them for following natural language instructions~\cite{hemachandra2015learning, patki2019inferring, blukis2018following, blukis2019learning, anderson2019chasing, patki2020language} has a long history in robotics.
Common approaches can be classified into sparse topological and dense grid-based maps.

\citet{walter2013learning} introduced a sparse semantic graph that combines pose, semantic, and topological information, extracted from sensory observations and speech descriptions along a route.
\citet{hemachandra2014learning} added a spatial map layer, and fused language with other sensory modalities.
\citet{hemachandra2015learning} used these representations for grounding natural language route instructions.
More recently, \citet{patki2019inferring} extended this framework to build compact world models specific to the input instruction, and \citet{patki2020language} enabled supporting previously unseen environments.
This class of sparse topological maps are well suited for probabilistic language grounding from symbolic representations.

Dense grid-based 2D semantic maps are suited for downstream processing using learned neural network modules, and have been used in modular neural network approaches for language grounding~\cite{gordon2017iqa, blukis2018mapping, blukis2019learning, anderson2019chasing, saha2021modular}.
\citet{saha2021modular} used a grid-based spatial representation and a map filtering method, showing promising early results on a subset of the \alfred dataset.
We extend this line of work to 3D voxel maps, add explicit tracking of occupancy and observability, and maintain the representation through time to facilitate grounding high-level language over long time horizons.
Our dense representation has a number of advantages.
First, it is easy to build in real-time from RGBD data using segmentation models and geometric operations.
Second, it captures structures found in indoor environments, such as L-shaped countertops or kitchen islands with sinks that are hard to represent topologically.
Third, it encodes spatial object relationships without requiring an ontology of spatial relations, or even tracking of object instances.
The main limitation of our approach is a memory footprint that scales with the physical size of the environment, making it less suited for outdoor or field applications.
Follow-up work could address this limitation, for example by using multi-scale representations such as octrees~\cite{meagher1982geometric, takikawa2021neural}.

\paragraph{Detailed comparison to HiTUT}
\label{app:comp:hitut}

We provide a detailed technical comparison  between our approach and HiTUT~\cite[Hierarchical Task Learning from Language Instructions with Unified Transformers and Self-Monitoring;][]{zhang2021hitut}, our main point of comparison.

Both our approach and HiTUT use a hierarchical task decomposition of goals into sequences of subgoals, and subgoals into sequences of actions.
The set of subgoals assumed by our approach and HiTUT have differences.
HiTUT has an additional subgoal \textsc{GoTo}(\textsc{Location}), while we view any navigation as a means to an end of a manipulation subgoal, and therefore do not have an explicit \textsc{GoTo} subgoal.
HiTUT additionally has subgoals for \textsc{Clean} and \textsc{Heat}, (e.g., \textsc{Clean}(\textsc{Obj}) usually abstracts over the sequence \textsc{Put}(\textsc{Sink}), \textsc{ToggleOn}(\textsc{Faucet}), \textsc{ToggleOff}(\textsc{Faucet}), \textsc{Pickup}(\textsc{Obj})), while our high-level policy would have to predict this entire sequence.

In terms of the model architecture, we use a hierarchical model with high-level and low-level controllers to mimic the task structure.
In contrast, HiTUT uses a flat transformer model to jointly solve high-level subgoal planning and low-level action prediction.
One of their main contributions is showing how a flat transformer model can be used to model a hierarchical task structure.
The benefit of our hierarchical model decomposition in combination with a shared spatial state representation is its ability to solve low-level navigation and manipulation problems with specialized modules, while avoiding the representational error accumulation and representation engineering issues typically associated with modular pipeline approaches. 

In terms of inference, HiTUT and our approach both sample subgoals one at a time, dynamically responding to changes in environment and execution. Both approaches perform backtracking to previous subgoals upon subgoal failure.

In terms of perception, our approach requires a pre-trained segmentation model, while HiTUT requires a pre-trained object detection model to generate object entity information that is fed into the transformer.

\subsection{Observation Model Details}\label{app:obsmodel}

\begin{figure*}[h!]
\centering
\includegraphics[width=\textwidth]{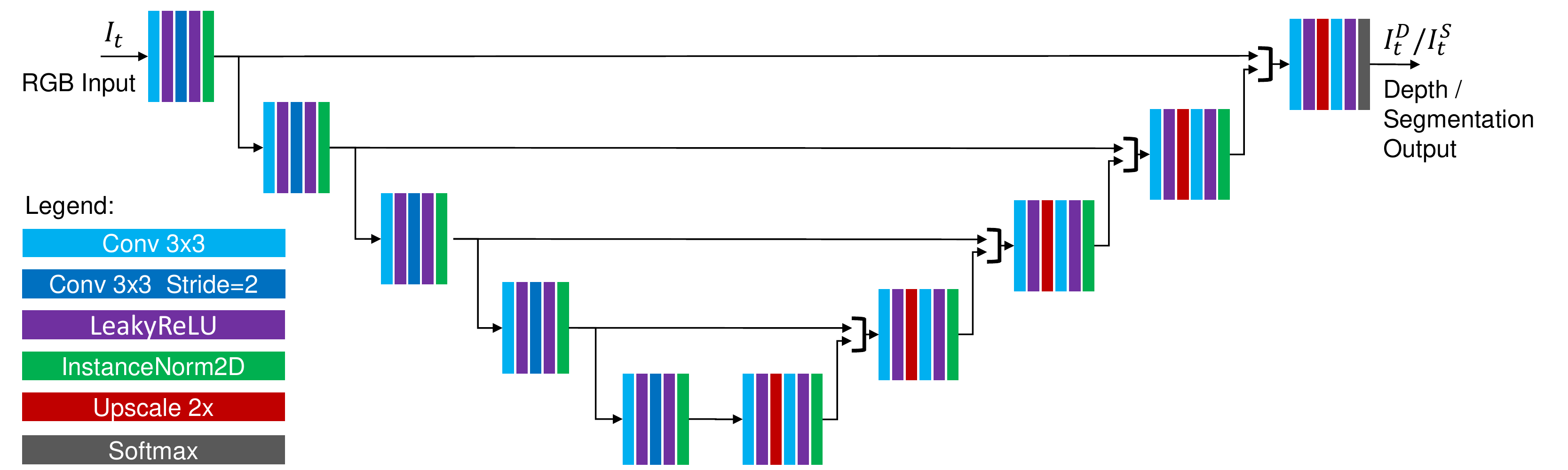}
\caption{Illustration of the U-Net architecture used in the depth and segmentation networks.}
\label{app:fig:unet}
\end{figure*}

At time $t$, during the perception step, we predict first-person semantic segmentation $\egosegmentation_{t}$ and depth $\depth_{t}$ from the observation $\obs_{t} = (\image_{t}, \pose_{t}, \inventory_{t}, \taskstr)$, from the RGB image $\image_{t}$ with neural network models pre-trained in the \alfred environment.
Each pixel $[\egosegmentation_{t}]_{(u,v)}$ at coordinates $(u,v)$ is a distribution over $C$ object classes.
Likewise, $[\depth_{t}]_{(u,v)}$ is a distribution over $B$ uniformly spaced depth bins $\{0, \Delta_{D}, 2\Delta_{D}, \dots, (B-1)\Delta_{D} \}$, where $\Delta_{D}$ is a depth resolution. 
In early experiments, we observed that  $\Delta_{D}$ should be less than 50\% of the voxel size.
We use $\Delta_{D} = 0.1m$, $B = 50$, and voxel size of $0.25m$.
We also heuristically compute a binary mask $M^{D}_{t}$ that indicates which pixels have confident depth readings. We allow more confidence slack in pixels that correspond to the current subgoal argument $\hlactionarg_{t}$ according to $\egosegmentation_{t}$.
The mask $M^{D}_t$ is used in the projection step to discard points $(x,y,z)$ that correspond to pixels $(u,v)$ for which $[M^{D}]_{(u,v)} = 0$.
The mask computation is: 
\begin{equation}
M^{D}_{t} = (W_{90}[\depth_{t}] < c_1 \E[\depth_{t}]) \lor ((W_{90}[\depth_{t}] < c_2 \E[\depth_{t}]) \land ([\egosegmentation_{t}]_{\hlactionarg_{t}} > 0.5))\;\;,
\end{equation}
where $W_{90}[\depth_{t}]$ is the width of the 90\% confidence interval at each pixel,
$\E[\depth_{t}]$ is the expected depth at each pixel, 
and $[\egosegmentation_{t}]_{\hlactionarg_{t}}$ is a 0-1 valued segmentation mask of the class of the current subgoal argument.
We set the hyperparameters  $c_1 = 0.3$ and $c_2 = 1.0$ to allow higher depth uncertainty for points corresponding to the subgoal argument.

If the agent is currently holding an object (i.e. $\sum_i[[\inventory_{t}]_{(i)}]$ > 0), we also discard points closer than $0.7m$ to the camera to make sure that the object in the agent inventory does not get added to the voxel map.

We use custom models based on the U-Net architecture~\cite{ronneberger2015unet} for depth and segmentation networks.
The architecture is illustrated in Figure~\ref{app:fig:unet}.
It consists of a cascade of five downscale blocks followed by five upscale blocks with skip-connections.
Each block includes two convolutions, two leakyReLU activations, and an instance normalization layer. The upscale blocks contain a 2x spatial upscaling operation.
We found that training a separate network for depth and segmentation worked better than sharing one network for both tasks.

\subsection{Model Execution Flow}
\label{app:exec:flow}

\begin{algorithm}
\begin{algorithmic}[1]
\footnotesize
\Require Instr. $L$, Horizon $T_{max}$.
\State $g_{0,1,2 \dots}, \staterepr_{-1} \leftarrow \texttt{null}$
\State $k \leftarrow 1$
\State Observe initial $\obs_0$.
\For{$t = 0, 1, 2, \dots H$}
    \State $\staterepr_{t} \leftarrow \obsfunc(\staterepr_{t-1}, \obs_t, \hlaction_k)$
    \label{algo:exec:obsfunc}
    \Do
        \If{$\hlaction_{k} = \texttt{null}$}
            \State $\hlaction_k \sim \hlpolicy(\taskstr, \staterepr_t, \langle \hlaction_i \rangle_{i < k})$
            \label{algo:exec:hlc}
            \If{$\hlaction_{k} = \stopsubgoal$}
                \State End episode
                 \label{algo:exec:end_stop}
            \EndIf
        \EndIf
        \State $\action_t \sim \llpolicy(\hlaction_k, \staterepr_t)$
        \label{algo:exec:llc}
        \If{$\action_t = \passaction$}
            \State $k \leftarrow k + 1$
            \label{algo:exec:pass}
        \EndIf
        \If{$\action_t = \failaction$}
            \State $\hlaction_k \leftarrow \texttt{null}$
            \label{algo:exec:fail}
        \EndIf
    \doWhile{$\action_t \in \{\failaction, \passaction\}$}
    \State Perform $\action_t$, observe $\obs_{t+1}$
    \label{algo:exec:exec}
\EndFor
\State End episode
\label{algo:exec:end_h}
\end{algorithmic}

\captionof{algorithm}{Execution Flow}
\label{algo:exec}
\end{algorithm}

Algorithm~\ref{algo:exec} describes the execution flow.
At time $t=0$ the initial observation $\obs_0$ is received.
At each timesep, we update the state representation $\staterepr_t$ (Line~\ref{algo:exec:obsfunc}).
If needed, we sample a new subgoal $\hlaction_k$ from $\hlpolicy$ (Line~\ref{algo:exec:hlc}), and then sample an action $\action_t$ from $\llpolicy$.
If $\action_t$ is $\passaction$, we increment subgoal counter $k$ (Line~\ref{algo:exec:pass}).
If it is $\failaction$, we discard the current subgoal $k$ (Line~\ref{algo:exec:fail}).
We repeat Lines~\ref{algo:exec:hlc}--\ref{algo:exec:fail} until an executable action $\action_t$ is sampled.
We execute $\action_t$, receive the next observation (Line~\ref{algo:exec:exec}), and proceed to the next timestep.
The episode ends when the subgoal $\stopsubgoal$ is sampled (Line~\ref{algo:exec:end_stop}) or the horizon $T_{max}$ is exceeded (Line~\ref{algo:exec:end_h}).

\subsection{High-Level Controller Details}

\begin{figure*}[t!]
\centering
\includegraphics[width=\textwidth]{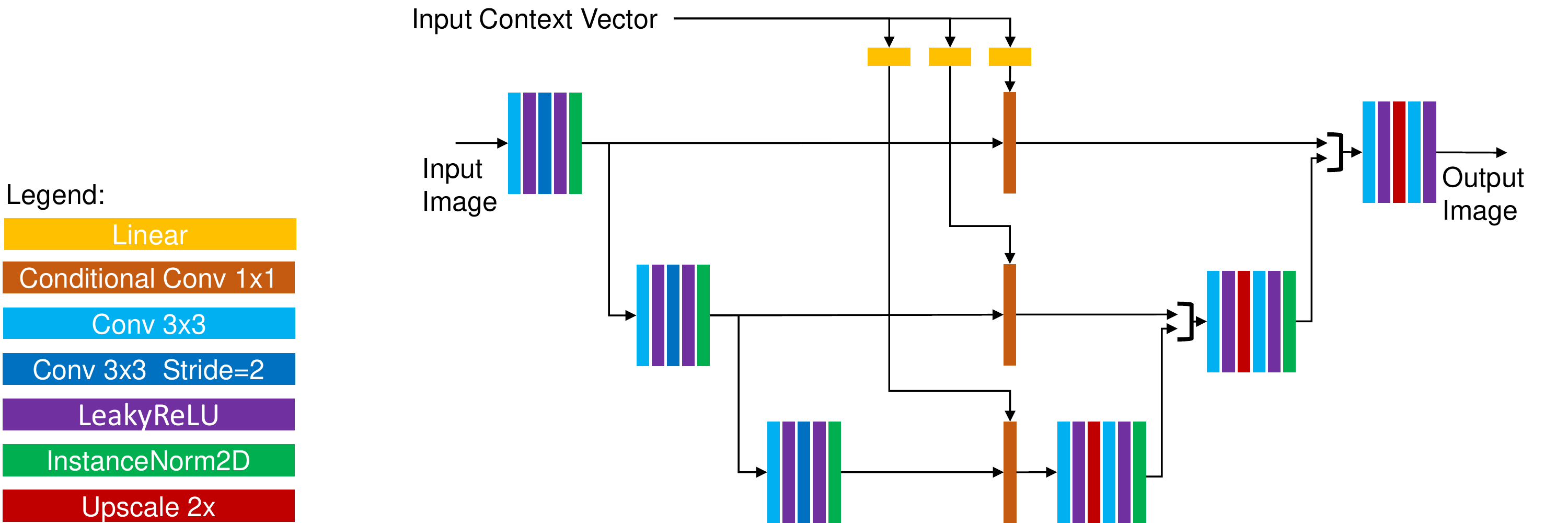}
\caption{Illustration of the LingUNet architecture used for as part of $\maskrefiner$ within the high-level controller $\hlpolicy$, and as part of the navigation model $\navmodel$ within the low-level controller.
The conditional convolutions parameters are computed during the network forward pass.}
\label{app:fig:lingunet}
\end{figure*}

Subgoals are predicted periodically. 
Let $\hlaction_k = (\hlactiontype_k, \hlactionarg_k, \hlactionmask_k)$ be the $k$-th subgoal predicted at time $t$. 
Predicting  the subgoal type $\hlactiontype_k$ and the argument class $\hlactionarg_k$ is described in the main paper (Section~\ref{sec:model:hlp}).
This section provides further details of  $\maskrefiner$, the model we use to generate $\hlactionmask_k$.
The mask refiner $\maskrefiner$ has four inputs:\footnote{Errata: Equation~\ref{eq:refiner} in the main paper is missing $[\semanticgrid_t]_{(\hlactionarg_k)}$ and $\pose_t$ arguments to the $\maskrefiner$.} (a) a spatial feature map $\mathbf{x}^{ego}_t \in [0,1]^{N \times W \times L}$ oriented in the agent egocentric reference frame; (b) $[\semanticgrid_t]_{(\hlactionarg_k)} \in [0,1]^{W \times L \times H}$, a 3D mask indicating all voxels that contain objects of class $\hlactionarg_k$ in the voxel map $\semanticgrid_t$; (c) the agent's pose $\pose_t$; and (d) a vector representation of the instruction $\taskemb$.
It outputs a 3D mask $\hlactionmask_k \in [0,1]^{W \times L \times H}$ that identifies the subgoal argument object. Formally, the computation is:\footnote{$\otimes$ is an operation that multiplies a $W \times L$ matrix by a $W \times L \times H$ tensor to obtain a $W \times L \times H$ tensor} 
\begin{align}
    \maskrefiner(\mathbf{x}^{ego}_t, [\semanticgrid_t]_{(\hlactionarg_k)}, \pose_t, \taskemb) = & \\ & \hspace{-7em} \textsc{AlloTransform}(\lingunet_m(\mathbf{x}^{ego}_t, \taskemb), \pose_t) \otimes [\semanticgrid_t]_{(\hlactionarg_k)}\;\;,\nonumber
\end{align}
where $\textsc{AlloTransform}$ transforms a spatial 2D map from an egocentric to the global reference frame, and $\lingunet_m$ is a language-conditioned image-to-image encoder-decoder~\cite{misra2018mapping}.
The architecture of $\lingunet_m$ is illustrated in Figure~\ref{app:fig:lingunet}.

\subsection{Low-Level Controller Details}\label{app:model:llp}

We describe the implementation of each of the low-level controller procedures. 
This implementation is not the focus of this paper, and could be improved or replaced with other algorithms. 
Some of the procedures cause actions in the AI2Thor environment, others simply process data to pass between procedures. 

\begin{figure*}[t!]
\centering
\includegraphics[width=\textwidth]{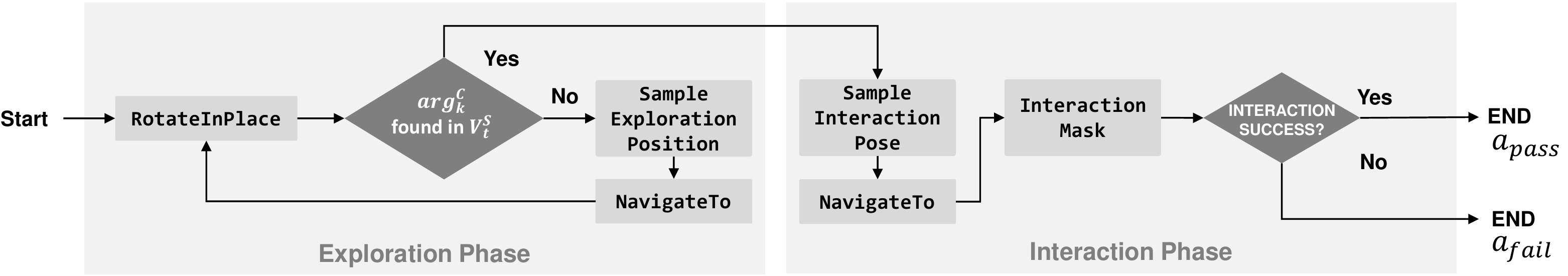}
\caption{Illustration of the low-level controller execution flow, showing the order in which procedures are used to complete a subgoal $\hlaction_k$.}
\label{app:fig:llcflow}
\end{figure*}

The procedures are \texttt{NavigateTo}, \texttt{SampleExplorationPosition}, \texttt{SampleInteractionPose}, and \texttt{InteractMask}.
The low-level controller receives the subgoal $\hlaction_k$, and follows a pre-specified execution flow across multiple timesteps to complete it.
The execution flow (Figure~\ref{app:fig:llcflow}) consists of an \emph{exploration} and \emph{interaction} phase.
In the exploration phase, we perform a 360\textdegree{} rotation by generating a sequence of three $\textsc{RotateLeft}$ actions to observe the environment and add information to the semantic map.
If the semantic map indicates that no object of type $\hlactionarg_k$, the action argument, is observed, we explore the environment by sampling a position $(x, y) = \texttt{SampleExplorationPosition}(\staterepr_{t})$, navigating there using  $\texttt{NavigateTo}(x,y,\staterepr_{t})$, and performing another 360\textdegree{} rotation.
We repeat this process until a voxel in $\semanticgrid_{t}$ contains the class $\hlactionarg_k$ with $>$50\% probability, at which point we move on to the interaction phase.
In the interaction phase, we sample an interaction pose $(x, y, \yaw, \pitch) = \texttt{SampleInteractionPose}(\staterepr_{t}, \hlaction_{k})$, invoke $\texttt{NavigateTo}(x, y, \staterepr_{t})$ to reach the position (x,y), and rotate according to yaw and pitch angles $(\yaw, \pitch)$.
Finally, we generate the egocentric interaction action mask $\actionmask_{t} = \texttt{InteractMask}(\staterepr_{t}, \hlactionmask_k)$, and execute the interaction action $(\hlactiontype_k, \actionmask_{t})$ in the \alfred environment.
We output $\passaction$ or $\failaction$ depending if the interaction action has succeeded, and pass control back to the high-level controller to sample the next subgoal.

\subsubsection{NavigateTo Procedure}
\newcommand{\vinstates}{\mathcal{S}^{vin}}
\newcommand{\vinstate}{s^{vin}}
\newcommand{\vinactions}{\mathcal{A}^{vin}}
\newcommand{\vinaction}{a^{vin}}
\newcommand{\vintransition}{\mathcal{T}^{vin}}
\newcommand{\vinreward}{R^{vin}}
\newcommand{\vinterm}{\mathcal{M}^{vin}}
\newcommand{\vinwidth}{W^{vin}}
\newcommand{\vinheight}{H^{vin}}
\newcommand{\vinq}{Q^{vin}}

At time $t$, the \texttt{NavigateTo} procedure maps a 2D navigation goal position $(x,y)$ and the state representation $\staterepr_{t}$ to one of the actions: $\{\rotateleft, \rotateright, \moveahead, \stopaction\}$.
We implement it with a Value Iteration Network~\cite[VIN;][]{tamar2016valueiter} that solves a 2D grid-MDP to predict navigation actions using fast GPU-accelerated convolution and max-pooling operations. The VIN parameters are pre-defined, and not learned. 
Other motion planners such as A$^{*}$ could be used as well.

The VIN is defined by
a state-space $\vinstates$,
action space $\vinactions$,
transition function $\vintransition : \vinstates \times \vinactions \rightarrow \vinstates$, 
a reward function $\vinreward : \vinstates \times \vinactions \rightarrow \vinactions$, 
and terminal state set $\vinterm$.
At each timestep $t$, VIN performs value iteration to compute the Q-function $\vinq : \vinstates \times \vinactions \rightarrow \mathbbm{R}$ that estimates the expected sum of future discounted rewards for taking action $\vinaction_t \in \vinactions$ in state $\vinstate_t \in \vinstates$, and thereafter following a greedy policy: $\vinaction_i = \arg \max_{\vinaction \in \vinactions} Q(\vinstate_i, \vinaction)$, $i > t$.
We implement the state space $\vinstates$ as a 2D grid of shape $\vinwidth \times \vinheight$.
Each state $\vinstate$ is tagged with three 0-1 valued attributes: $\textsc{Obstacle}, \textsc{Unobserved}, \textsc{Goal}$.
At each timestep $t$, we set the values of state attributes according to the most recent state representation $\staterepr_t$ and current navigation goal $(x,y)$.
States $\vinstate$ with occupied voxels in the height range $[0, 1.75m]$ are tagged $\textsc{Obstacle}(\vinstate) = 1$, otherwise $\textsc{Obstacle}(\vinstate) = 0$.
States with all voxels unobserved are tagged $\textsc{Unobserved}(\vinstate) = 1$, otherwise $\textsc{Unobserved}(\vinstate) = 0$.
The state at the goal position is tagged $\textsc{Goal}(\vinstate) = 1$, for all others $\textsc{Goal}(\vinstate) = 0$.
The action space is $\vinactions = \{\textsc{North}, \textsc{East}, \textsc{South}, \textsc{West}, \textsc{Stop}\}$.
The transition function encodes epsilon-greedy grid navigation dynamics: (1) the action $\textsc{North}$ moves the agent one state north and likewise for other actions, and (2) with probability $\epsilon = 8\%$ a random transition to a neighboring state occurs.
Visiting any terminal state $\vinstate \in \vinterm$ or executing the action $\textsc{Stop}$ terminates the episode.
Terminal states are all states tagged with attributes $\textsc{Obstacle}$ and $\textsc{Goal}$, $\vinterm = \{\vinstate \in \vinstates \; \mid \; (\textsc{Obstacle}(\vinstate) > 0.5) \land (\textsc{Goal}(\vinstate) > 0.5)\}$.

The reward function assigns different rewards for visiting states with different attributes:
\begin{multline}
    \vinreward(\vinstate, \vinaction) = -0.9 \cdot \textsc{Obstacle}(\vinstate) + 1.0 \cdot \textsc{Goal}(\vinstate) 
    \\- 0.02 \cdot \textsc{Unobserved}(\vinstate) + 0.001 \cdot \mathbbm{1}_{\vinaction = \textsc{Stop}}\;\;.
\end{multline}
$\textsc{Obstacle}$ states receive reward $-0.9$, $\textsc{Goal}$ states receive reward $1.0$, and $\textsc{Unobserved}$ states receive reward $-0.02$.
Taking the $\textsc{Stop}$ action in any state gives reward $0.001$, which has the effect of the agent stopping in unsolvable cases.
We use the VIN iteratively for $N^{vin}$ iterations, and predict an action $\vinaction = \arg \max_{\vinaction \in \vinactions}(\vinq(\vinstate_t, \vinaction)$.
We map from the VIN action $\vinaction$ to a single valid AI2Thor navigation action using a deterministic mapping (Table~\ref{app:tab:actionmapping}).

\begin{table}[t]
\ra{1.1}
    \centering
    \footnotesize
    \begin{tabular}{ccc}
        \toprule
        \textbf{Current Heading} & \textbf{VIN Action} & \textbf{AI2Thor Action} \\
        \midrule
         & $\textsc{West}$ & $\rotateleft$ \\
        North & $\textsc{North}$ & $\moveahead$ \\
         & $\textsc{East}$ or $\textsc{South}$ & $\rotateright$ \\
         \midrule
        & $\textsc{North}$ & $\rotateleft$ \\
        East & $\textsc{East}$ & $\moveahead$ \\
         & $\textsc{South}$ or $\textsc{West}$ & $\rotateright$ \\
         \midrule
        & $\textsc{East}$ & $\rotateleft$ \\
        South & $\textsc{South}$ & $\moveahead$ \\
         & $\textsc{West}$ or $\textsc{North}$ & $\rotateright$ \\
         \midrule
                 & $\textsc{South}$ & $\rotateleft$ \\
        West & $\textsc{West}$ & $\moveahead$ \\
         & $\textsc{North}$ or $\textsc{East}$ & $\rotateright$ \\
        \bottomrule
    \end{tabular}
    \vskip 8 pt
    \caption{Mapping from VIN actions to AI2Thor actions.}
    \label{app:tab:actionmapping}
\end{table}

\subsubsection{SampleExplorationPosition}

The \texttt{SampleExplorationPosition} procedure maps a state representation $\staterepr_t$ to a discrete 2D position $p^{\rm explore} = (x,y)$.
Let $\mathcal{P}_s$ be the set of 2D positions corresponding to voxel centroids in the voxel map along the horizontal axes, and the ground set $\mathcal{P}_{g}$ as the set of all unoccupied positions that have the class $\textsc{Floor}$ or $\textsc{Rug}$ in at least one voxel.
A position is unoccupied if all voxels in the height range $[0, 1.75m]$ are free of obstacles.
We define a frontier set $\mathcal{P}_{f}$ as the set of all positions $\mathcal{P}_g$ for which at least one immediately neighboring position contains zero observed voxels.
If $\mathcal{P}_f$ is non-empty, we sample the position $p^{\rm explore}$ uniformly at random from $\mathcal{P}_f$.
Otherwise, we sample $p^{\rm explore}$ uniformly at random from $\mathcal{P}_g$.

\subsubsection{SampleInteractionPose}

The \texttt{SampleInteractionPose} procedure maps the state representation $\staterepr_t$ and subgoal $\hlaction_k = (\hlactiontype_k, \hlactionarg_k, \hlactionmask_k)$ to a pose $\pose =  (x, y, \yaw, \pitch)$, where $(x,y)$ is a discrete 2D position, $\yaw$ is the agent yaw angle, and $\pitch$ is the agent camera pitch angle.
The pose is predicted such that upon reaching it, the interaction action of type $\hlactiontype_k$ is likely to succeed on the object of class $\hlactionarg_k$ at location identified by the mask $\hlactionmask_k$.

We use a neural network model $\navmodel$ to predict expected pitch $\mathbbm{E}(\pitch | x, y; \hlaction_k, \staterepr_t)$ and a distribution $P(x, y, \yaw | \hlaction_k, \staterepr_t)$, factored as:
\begin{equation}
    P(x, y, \yaw | \hlaction_k, \staterepr_t) = 
    P(\yaw | x, y; \hlaction_k, \staterepr_t)P(x, y | \hlaction_k, \staterepr_t)
\end{equation}
The network $\navmodel$ is based on the LingUNet architecture (Figure~\ref{app:fig:lingunet}):
\begin{multline}
    \navmodel(\staterepr_t, \hlaction_k) = \\ \lingunet(\featurize(\staterepr_t), \textsc{Linear}([\textsc{Lut}_T(\hlactiontype_k); \textsc{Lut}_C(\hlactionarg_k)]))\;\;,
\end{multline}
where $\featurize$ is an affordance feature map (Section~\ref{sec:model:staterepr}), $\textsc{Linear}$ is a linear layer with bias, $\textsc{Lut}_T$ and $\textsc{Lut}_C$ are embedding lookup tables, and $[\cdot ; \cdot]$ is a vector concatenation.

To sample a pose $\pose$, we first sample a position $(x,y) \sim P(x, y | \hlaction_k, \staterepr_t)$, then sample a yaw angle $\yaw \sim P(\yaw | x, y; \hlaction_k, \staterepr_t)$, and finally lookup a pitch angle $\pitch = \mathbbm{E}(\pitch | x, y; \hlaction_k, \staterepr_t)$.

\subsubsection{InteractionMask}

The \texttt{InteractionMask} procedure maps a state representation $\staterepr_t = (\semanticgrid_t, \observedgrid_t, \inventory_t, \pose_t)$, the most recent RGB observation $\image_t$, the most recent predicted segmentation $\egosegmentation_t$, and a subgoal $\hlaction_k = (\hlactiontype_k, \hlactionarg_k, \hlactionmask_k)$ to a 0-1 valued mask $\actionmask_t \in [0,1]^{H \times W}$ that identifies the interaction object in the first-person view observation.
The interaction mask $\actionmask_t$ is in the format expected by \alfred. 
Formally, it is computed in three steps:
\begin{align}
    \actionmask_t^{A} &= [\egosegmentation_t]_{\hlactionarg_k}\\
    \actionmask_t^{B} &= \textsc{PinholeCam}(\hlactionmask_k, \pose_t)\\
    \actionmask_t &= \actionmask_t^{A} \cdot \actionmask_t^{B}\;\;,
\end{align}
where $\textsc{PinholeCam}$ projects the 0-1 valued 3D voxel map $\hlactionmask_k$ to the agent's camera plane according to the pose $\pose_t$.
The mask $\actionmask_t^{A}$ is an egocentric 0-1 valued mask that identifies all objects of class $\hlactionarg_k$ in the image $\image_t$.
The $\actionmask_t^{B}$ is an egocentric 0-1 valued mask that identifies the voxels $\hlactionmask_k$.
For each pixel $(u,v)$, the value $[\actionmask_t^{B}]_{(u,v)}$ is the maximum of all values $[\hlactionmask_k]_{(x,y,z)}$ over voxels with coordinates $(x,y,z)$ that the ray cast from the camera through the pixel $(u,v)$ intersects with.
The final mask $\actionmask_t$ is a 0-1 valued mask that identifies not only the correct object class, but also the correct instance according to the voxel mask $\hlactionmask_k$.

\subsection{Additional Learning Details}\label{app:learning}

\subsubsection{Observation Model Learning}\label{app:learning:obs}

\paragraph{Data}
As described in Section~\ref{sec:learning}, we use a perception dataset $\mathcal{D}^{P}$ for training depth and segmentation models.
The dataset $\mathcal{D}^{P} = \{([\image]^{(i)}, [\depth]^{(i)}, [\egosegmentation]^{(i)}\}_{i=1}^{N_{P}}$ includes RGB images $[\image]^{(i)}$ with ground truth depth $[\depth]^{(i)}$ and segmentation $[\egosegmentation]^{(i)}$.
The ground truth depth $[\depth]^{(i)}$ at each pixel $(u,v)$ is a distribution $[\depth]^{(i)}_{((u,v))}$ over $B$ depth bins, where 100\% of the probability mass is assigned to the bin containing the reference depth value.
The ground truth segmentation $[\egosegmentation]^{(i)}$ is likewise at each pixel $(u,v)$ a one-hot vector indicating the object class that pixel belongs to.

\paragraph{Data Augmentation}
The \alfred dataset consists of 108 different training scenes, where each scene has a fixed furniture and light fixtures.
Observations are highly correlated within each scene, which greatly reduces the effective size of the perception dataset and hurts generalization to unseen scenes.
We use a custom segmentation-aware data augmentation strategy that increases the diversity of RGB observations.

\begin{figure*}[t]
\centering
\includegraphics[width=\textwidth]{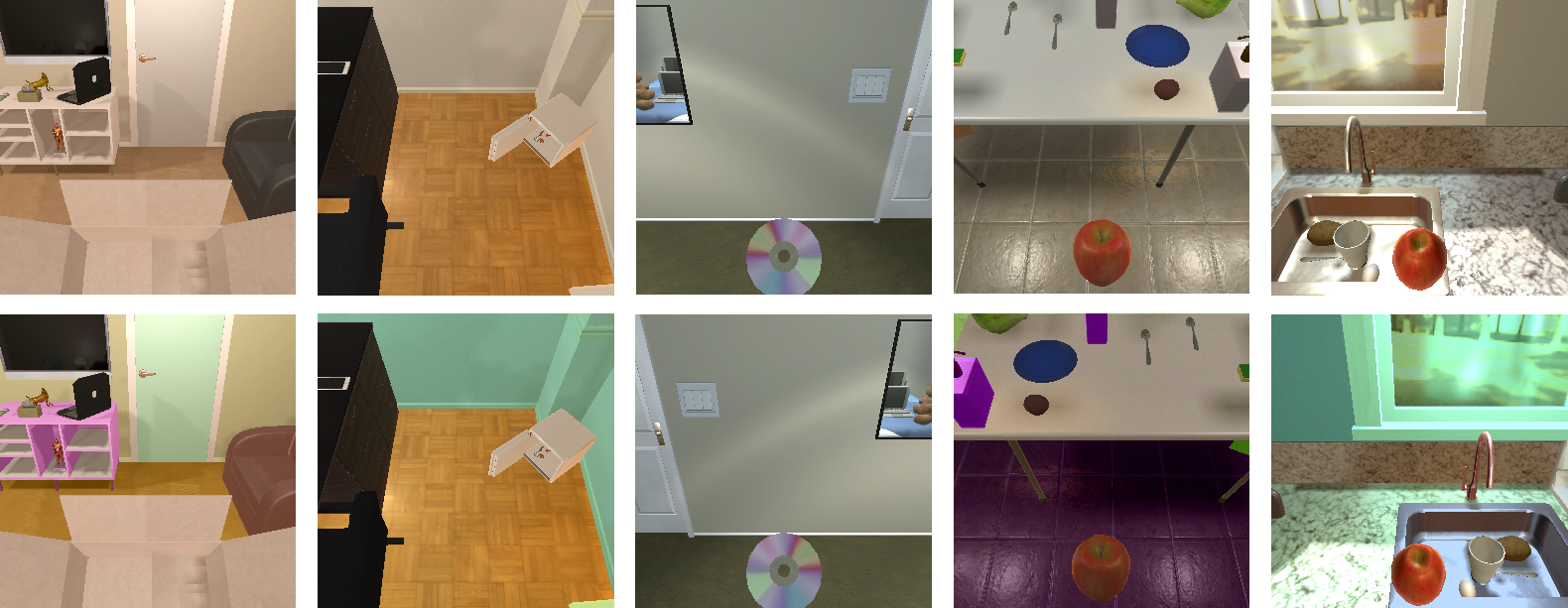}
\caption{Examples of images produced with our $\textsc{Augment}$ procedure. The top row shows raw RGB images from \alfred. The bottom row shows images generated by our  segmentation-aware data augmentation method.
Objects like walls, sinks, floors, and furniture randomly change color, while the apple and the spoons do not.}
\label{app:fig:augment}
\end{figure*}

We compute an augmented RGB image $\tilde{\image} = \textsc{Augment}(\image, \egosegmentation, O_{v})$ that maps the image $\image$, ground-truth segmentation $\egosegmentation$, and a set of semantic classes $O_{v}$ to a new image $\tilde{\image}$.
$O_{v}$ is the set of object classes that are likely to appear in different colors.
$O_{v}$ includes classes like \textsc{Floor}, \textsc{CounterTop}, \textsc{Cabinet}, \textsc{Vase}, \textsc{SoapBottle}, \textsc{AlarmClock} that come in different designs and colors, but not classes like \textsc{Banana}, \textsc{Apple}, \textsc{Spoon} that tend to have even appearance.
Algorithm~\ref{app:algo:augment} shows the implementation of $\textsc{Augment}$.
It emulates more diverse environments by applying a different random color offset to each object class in the RGB image.
Figure~\ref{app:fig:augment} shows examples of images produced with this augmentation procedure.

During training, we apply $\textsc{Augment}$ with 50\% probability to each training example.
Additionally, with 50\% probability we perform a horizontal flip.

\begin{algorithm}[t]
\caption{$\textsc{Augment}$}
\begin{algorithmic}[1]
\footnotesize
\Require RGB Image $\image$, ground truth segmentation $\egosegmentation$, set of object classes $O_{v}$.
\State $\tilde{\image} \leftarrow \image$
\For{$c \in O_{v}$}
    \State \Comment{Extract a binary mask corresponding to object class $c$}
    \State $M_c \leftarrow [\egosegmentation]_{(c)}$

    \State \Comment{Apply modifications to the image, masked by the segmentation mask $M_c$}
    \State \Comment{\quad $\odot$ multiplies a $N$-dimensional vector with a $H \times W$ tensor to compute a $N \times H \times W$ tensor}
    \State \Comment{\quad $\cdot$ is an elementwise multiplication}

    \If{$\texttt{randomBernoulli}(0.5)$}
        \State \Comment{Sample an additive color offset for class $c$ from a normal distribution.}
        \State \Comment{\quad $I_3$ is the $3\times 3$ identity matrix.}
        \State $a \sim N(\overrightarrow{\mathbf{0}}, \sigma_a I_{3})$
        \State $\tilde{\image} \leftarrow \tilde{\image} + a \odot M_c$
    \EndIf
    \If{$\texttt{randomBernoulli}(0.5)$}
        \State \Comment{Sample additive gaussian noise for each pixel $(u,v)$ for class $c$}
        \For{each pixel $(u,v)$}
            \State $g_{u,v} \sim N(1, \sigma_g)$
            \State $[\tilde{\image}]_{(u,v)} \leftarrow[\tilde{\image}]_{(u,v)} + g_{u,v} \cdot [M_c]_{(u,v)}$
        \EndFor
    \EndIf
    \If{$\texttt{randomBernoulli}(0.5)$}
        \State \Comment{Sample an multiplicative color offset for class $c$ from a normal distribution}
        \State $m \sim N(\overrightarrow{\mathbf{0}}, \sigma_m I_{3})$
        \State $\tilde{\image} \leftarrow \tilde{\image} \cdot (m \odot M_c)$
    \EndIf
\EndFor
\State clamp image $\tilde{\image}$ within 0-1 bounds
\State \Return $\tilde{\image}$
\vspace{-3pt}
\end{algorithmic}
\label{app:algo:augment}
\end{algorithm}

\subsection{Additional Experimental Details}

We collect a training dataset of language-demonstration pairs as described in Section~\ref{sec:learning}.
The demonstrations in \alfred typically navigate while looking down at the floor, likely a side-effect of the PDDL planner that had access to the world state during data generation, and as such has no need to explore or observe the visual environment.
We modify the demonstration trajectories to get more informative first-person observations.
First, we insert four $\rotateleft$ actions at the start of each trajectory.
Second, we maintain a nominal camera pitch angle of 30\textdegree during navigation, by inserting $\lookdown$ and $\lookup$ actions before and after every interaction action.
We discard trajectories for which these modifications cause failures.
These modifications result in observations that are more useful for learning and constructing our persistent spatial representation.

\subsection{Hyperparameters}
Table~\ref{app:tab:hyperparams} shows hyperparameter values.
The hyperparameters were hand-tuned on the validation unseen split.

\subsection{Additional Results}
\label{app:results}

Additional qualitative results are available at: \url{https://hlsm-alfred.github.io/}.

A successful example of task execution is available at: \url{https://drive.google.com/file/d/1APKe3cR_-vliyU2elT5Un30w7PvkEdYs/view?usp=sharing}

A failed example of task execution is available at:
\url{https://drive.google.com/file/d/1j8BJ_ALoXGyf8a-IOkmQAg38awSWYt6f/view?usp=sharing}

\begin{table*}[t]
\ra{1.1} 
  \footnotesize
  \centering
  \begin{tabular}{cc}
  \toprule
  \textbf{Hyperparameter} & \textbf{Value} \\
  \midrule
  \multicolumn{2}{c}{\textbf{Observation Model}}\\
  \midrule
Number of depth bins $B$ & $50$\\
Depth resolution $\Delta D$ & $0.1m$\\
  \midrule
\multicolumn{2}{c}{\textbf{Spatial State Representation}} \\
  \midrule
Voxel Size & $0.25m$\\
Voxel Map Dimensions in Voxels & $61 \times 61 \times 10$ \\
Voxel Map Dimensions in Meters & $15.25m \times 15.25m \times 2.5m$ \\
Number of semantic classes $C$ & 123 \\

  \midrule
  \multicolumn{2}{c}{\textbf{High-level Controller}}\\
  \midrule
Subgoal history encoder hidden dimension & 128\\
Subgoal history encoder transformer layers & 2\\
Subgoal predictor dense MLP layers & 3\\
Subgoal predictor dense MLP hidden dimension & 128\\

  \midrule
  \multicolumn{2}{c}{\textbf{Low-level Controller}}\\
  \midrule
Number of VIN iterations $N^{vin}$ & $122$\\
VIN state space size & $61 \times 61$\\
  \midrule
  \multicolumn{2}{c}{\textbf{Development Environment}}\\
  \midrule
Programming Language & Python \\
ML and Math Library & PyTorch \\
  \bottomrule
  \end{tabular}
  \caption{Hyperparameter values.}
  \label{app:tab:hyperparams}
\end{table*}

\end{document}